%% file: main.tex
\documentclass[letterpaper]{article} % DO NOT CHANGE THIS
\usepackage[]{aaai23}  % DO NOT CHANGE THIS
\usepackage{times}  % DO NOT CHANGE THIS
\usepackage{helvet}  % DO NOT CHANGE THIS
\usepackage{courier}  % DO NOT CHANGE THIS
\usepackage[hyphens]{url}  % DO NOT CHANGE THIS
\usepackage{graphicx} % DO NOT CHANGE THIS
\usepackage{natbib}  % DO NOT CHANGE THIS AND DO NOT ADD ANY OPTIONS TO IT
\usepackage{caption} % DO NOT CHANGE THIS AND DO NOT ADD ANY OPTIONS TO IT
\usepackage{algorithm}
\usepackage{algorithmic}

\usepackage{multirow}
\usepackage[dvipsnames]{xcolor}
\usepackage[
raggedright
]{sidecap}
\usepackage{pifont}% http://ctan.org/pkg/pifont
\usepackage{amsmath}
\usepackage{amssymb}
\usepackage{tikz}

\usepackage{dashrule}
\usepackage{booktabs}
\makeatletter
\newcommand*\MohaDashedLine{\rotatebox[origin=c]{90}{$\dabar@\dabar@\dabar@\dabar@\dabar@\dabar@\dabar@\dabar@\dabar@\dabar@\dabar@\dabar@\dabar@\dabar@\dabar@\dabar@\dabar@\dabar@\dabar@\dabar@\dabar@\dabar@\dabar@\dabar@\dabar@\dabar@\dabar@\dabar@$}}
\makeatother
\makeatletter
\newcommand*\NicheDashedLine{\rotatebox[origin=c]{90}{$\dabar@\dabar@\dabar@\dabar@\dabar@\dabar@\dabar@\dabar@\dabar@\dabar@\dabar@\dabar@\dabar@\dabar@\dabar@\dabar@\dabar@\dabar@\dabar@\dabar@\dabar@\dabar@\dabar@\dabar@$}}
\makeatother

\newcommand{\cmark}{\ding{51}}%
\newcommand{\xmark}{\ding{55}}%

\usepackage{newfloat}
\usepackage{listings}
\DeclareCaptionStyle{ruled}{labelfont=normalfont,labelsep=colon,strut=off} % DO NOT CHANGE THIS
\floatstyle{ruled}
\newfloat{listing}{tb}{lst}{}
\floatname{listing}{Listing}
\title{Text-DIAE: A Self-Supervised Degradation Invariant Autoencoder for Text Recognition and Document Enhancement}
\author{
    Mohamed Ali Souibgui\equalcontrib\textsuperscript{\rm 1},
    Sanket Biswas\equalcontrib\textsuperscript{\rm 1},
    Andres Mafla\equalcontrib\textsuperscript{\rm 1},
    Ali Furkan Biten\equalcontrib\textsuperscript{\rm 1},
    Alicia Fornés\textsuperscript{\rm 1},
    Yousri Kessentini\textsuperscript{\rm 2},
    Josep Lladós\textsuperscript{\rm 1},
    Lluis Gomez\textsuperscript{\rm 1},
    Dimosthenis Karatzas\textsuperscript{\rm 1}
}
\affiliations{
    \textsuperscript{\rm 1} Computer Vision Center, Universitat Autònoma de Barcelona, Barcelona, Spain\\
    \{msouibgui, sbiswas, amafla, abiten, afornes, josep, lgomez, dimos\}@cvc.uab.es
    
    \textsuperscript{\rm 2} Digital Research Center of Sfax, SM@RTS Laboratory, Sfax, Tunisia\\
    yousri.kessentini@crns.rnrt.tn
    
}

\usepackage{bibentry}
\begin{document}

\maketitle

%  our files goes here
\begin{abstract}
\input{./tex/abstract.tex}
\end{abstract}

\section{Introduction}\label{s:intro}

\input{./tex/intro.tex}

\section{Related Work}\label{s:sota}
\input{./tex/sota.tex}

\section{Method}\label{s:method}
\input{./tex/method.tex}

\section{Experiments}\label{s:experiment}
\input{./tex/experiments.tex}

\section{Conclusion}\label{s:conclusion}
\input{./tex/conclusion.tex}
\section*{Acknowledgments}
\begin{small}
This work has been partially supported by  the Swedish Research Council (grant 2018-06074, DECRYPT), the Spanish projects RTI2018-095645-B-C21, CERCA Program / Generalitat de Catalunya, the FCT-19-15244,  the Catalan projects 2017-SGR-1783, PhD Scholarships from AGAUR (2021FIB-10010) and (2019-FIB01233), and from UAB (B18P0073). DocPRESERV project (Swedish STINT grant).
\end{small}

% Use \bibliography{yourbibfile} instead or the References section will not appear in your paper
\bibliography{main}

\clearpage

\input{appendix}

\end{document}

%% file: tex/abstract.tex
In this paper, we propose  a Text-Degradation Invariant Auto Encoder (Text-DIAE), a self-supervised model designed to tackle two tasks, text recognition (handwritten or scene-text) and document image enhancement. We start by employing a transformer-based architecture that incorporates three pretext tasks as learning objectives to be optimized during pre-training without the usage of labeled data. Each of the pretext objectives is specifically tailored for the final downstream tasks. We conduct several ablation experiments that confirm the design choice of the selected pretext tasks. Importantly, the proposed model does not exhibit limitations of previous state-of-the-art methods based on contrastive losses, while at the same time requiring \textit{substantially} fewer data samples to converge. Finally, we demonstrate that our method surpasses the state-of-the-art in existing supervised and self-supervised settings in handwritten and scene text recognition and document image enhancement. Our code and trained models will be made publicly available at~\url{ http://Upon_Acceptance}.

% \keywords{Self-Supervised Learning, Handwritten Text Recognition, Scene-Text Recognition, Document Image Enhancement.}

%% file: tex/intro.tex
In recent times, self-supervised learning paradigms have gained a lot of attention due to its ability of benefiting from massive unlabeled data which is easily accessible from different sources. However,  applying these approaches  remain quite limited in the domains of optical character recognition (OCR), handwritten text recognition (HTR) and document image enhancement, which motivate us to tackle the problem in this study.

Common computer vision pipelines using self-supervised frameworks employ a pretext-task (e.g. relative position prediction of patches~\cite{doersch2015unsupervised}, contrastive views~\cite{chen2020simple}, image inpainting~\cite{pathak2016context}, etc.) to learn visual representations for solving down-stream tasks like classification, object detection and so on. Current self-supervised paradigms~\cite{caron2021emerging, caron2020unsupervised, chen2020simple, chen2021empirical} have adapted transformers~\cite{vaswani2017attention} to learn visual representations from unlabeled images which are semantically meaningful. More recently, generative self-supervised approaches~\cite{he2021masked, bao2021beit, dong2021peco} using auto-encoders have been used to learn representations in the feature space through image patches and visual tokens.

Closely related to our work, some contributions in visual representation learning were addressing text recognition  {(HTR)~\cite{aberdam2021sequence,bhunia2021vectorization,liu2022perceiving} and Scene-Text Recognition (STR)~\cite{aberdam2021sequence,zhang2022context}) and image enhancement~\cite{liang2022semantically}}. Despite the performance gains, there are some drawbacks of such models: (1) independent sequences of tokens are treated as single data points, which can cause misalignment of similar sequences among a batch, (2) considerable batch size requirements to define negative contrastive pairs, (3) considerably slow convergence rates.

\begin{figure}[t]
\centering
\includegraphics[width=\columnwidth]{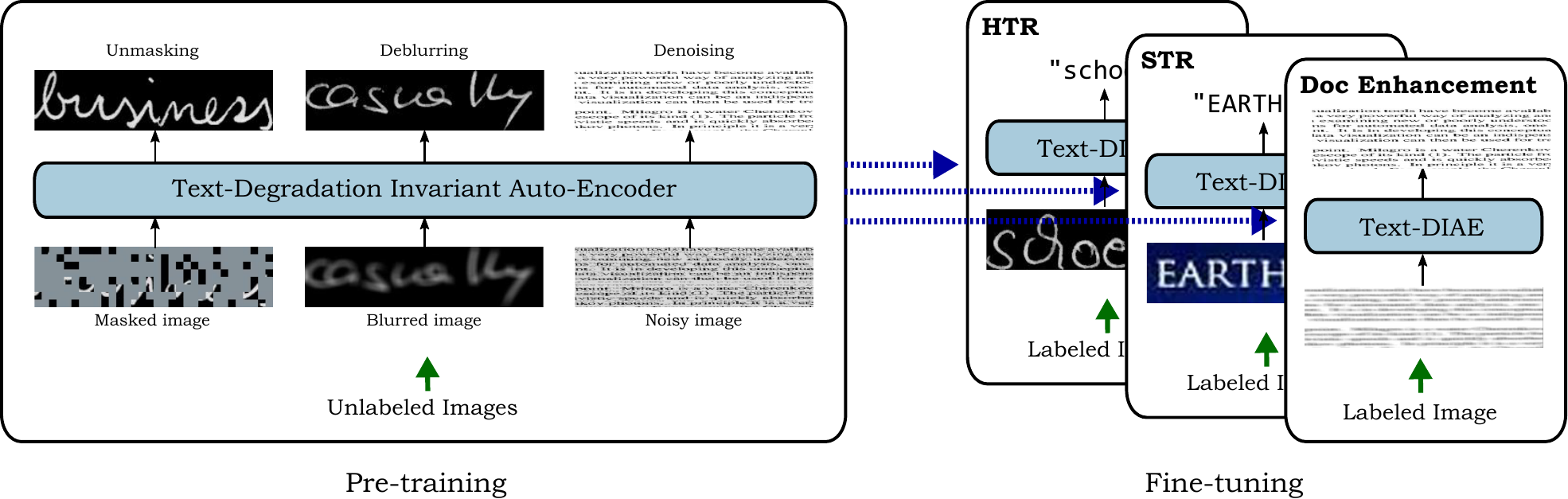}
\caption{\textbf{Text-Degradation Invariant Auto-Encoder (Text-DIAE),} we employ image reconstruction pretext tasks at pre-training. Masking, blurring and adding noise are employed to learn richer representations.}
\label{fig:teaser}
\end{figure}

For humans, reading text in noisy scenarios is possible because of the ability or reconstructing the degraded regions and predicting the missing/blurry content~\cite{howard1998anatomy, dehaene2014consciousness}. Incorporating such an ability in a model could immensely help in restoration, recognition and understanding of characters and symbols, considering that text carries rich linguistic information that allow humans to reason explicitly according to context. In order to endow this human-specific skill to our models,
% we adapt a self-supervised methodology that serves as a close proxy to two specific downstream tasks.
% Thus, 
we present in this paper a new self-supervised framework called Text-Degradation Invariant Auto-Encoders (Text-DIAE) inspired by the principle of denoising autoencoders~\cite{vincent2008extracting}, as depicted in Figure~\ref{fig:teaser}. 
Our model focuses on exploring the dynamics of learning representations under different degradation scenarios. Specifically, we propose the usage of a robust self-supervised auto-encoder along with customized pretext tasks (masking, blur and background noise) that are designed to specifically tackle two different downstream tasks: text recognition (HTR and STR) and document image enhancement (document binarization, document deblurring).
 {As a consequence, the choice of the proxy tasks have been realized to learn useful representations for solving these specific downstream tasks using unlabeled data}.

The benefits of employing such approach are: we do not define sequences at the feature level. Rather, by employing a transformer-based~\cite{vaswani2017attention} approach, similar to BERT~\cite{devlin2018bert} we utilize the self-attention layers to attend among patches which does not require big batches of negative samples. 
% This methodology simplifies the approach and avoids defining matching sequences per frames at a feature level. 
% Secondly, by employing masking, deblurring and enhancement as pretext tasks, our approach does not require big batches of negative samples, often required in contrastive learning approaches. 
Also, the combination of these pre-training tasks result in a significantly faster convergence compared to previous approaches. 
% Extensive experimentation is shown as ablation studies that measure the quality of the learned representations. We compare the learned capa and also compare them in semi-supervised (fine-tuning) settings. 
The resulting representations are evaluated by a scenario that resembles % The quality of the learned representations aims to resemble
the linear probing evaluation often used in self-supervision~\cite{kolesnikov2019revisiting, zhang2016colorful} and follows the scheme of~\cite{aberdam2021sequence} in text recognition task.
% The semi-supervised scenario fine-tunes the autoencoder with small percentages of labeled data. 
By this assessment, we find that our method outperforms previous self and semi supervised pipelines. 
Furthermore, by employing Text-DIAE, we achieve state-of-the-art in handwritten text recognition and document image enhancement, while outperforming scene text recognition under self-supervision settings.  {The essential findings and novelties of our work are based on the following interesting deductions: \begin{itemize}
    \item The impact and combination of pretext tasks depends on the downstream task.
    \item The closer the association between a pretext task and a downstream task, the better is the model performance.
    \item By employing Text-DIAE, we achieve faster convergence and use order of magnitude lesser data during pre-training than the contrastive-learning based approaches.
\end{itemize}  
To add on top of this, this is the first work to our knowledge that investigates different self-supervised pretext tasks for multiple significant downstream tasks in text recognition (HTR-word level, STR) and document image enhancement (document binarization, deblurring) while achieving state-of-the-art performance with {43} and {45} times lesser data for HTR and STR, respectively.}

%% file: tex/sota.tex
\noindent
\textbf{Self-Supervised Learning.}
Due to extensive efforts on labeled data requirements of supervised models, this learning paradigm emerges as a way of exploiting the structured information contained in data itself. 
Self-Supervised learning aims to obtain rich representations of an input modality by
% without explicitly incorporating the labels into the training procedure. 
% Such approaches have had a re-surge in computer vision  due to the recent success achieved in natural language processing~\cite{devlin2018bert, mikolov2013distributed, pennington2014glove, radford2019language, liu2019roberta} and speech recognition~\cite{baevski2020wav2vec, jiang2019improving, wang2020unsupervised}. 
% In order to acquire unlabeled signals coming from a given modality, 
designing pretext tasks that are used as auxiliary signals that are informative for a given downstream task. 
% In computer vision, 
Initial approaches relied on auto-encoders~\cite{vincent2008extracting} trained to remove artificially added noise from an image. Later, several approaches introduced other pretext tasks that provide rich signals to train a network as a feature extractor. Some pretext tasks employed were image colorization~\cite{zhang2016colorful}, jigsaw puzzle solving~\cite{noroozi2016unsupervised}, patch ordering~\cite{doersch2015unsupervised}, rotation prediction~\cite{gidaris2018unsupervised} among others.
Recent approaches rely on extensive image augmentation 
% transformations of a single input to a model that is trained 
to maximize the agreement among paired samples and contrast with all possible negative samples~\cite{chen2020simple, chen2020big, he2020momentum, zbontar2021barlow,  caron2020unsupervised, caron2021emerging}.

% ~\cite{chen2020simple, chen2020big, he2020momentum, hjelm2018learning, zbontar2021barlow, bachman2019learning, caron2020unsupervised, caron2021emerging, henaff2020data}. 

% Such approaches have rapidly escalated in performance and currently are on par with supervised methods. 
% More recently, an advance in representation learning have been achieved by the introduction of Masked Auto-encoders (MAE)~\cite{he2021masked}. The pretext task in a MAE is to predict a masked  latent representation of patches. Similar ideas have been also introduced~\cite{dong2021peco, bao2021beit}, which constantly show performance improvements in representation learning.
More recently, generative approaches like Masked Auto-encoders (MAE)~\cite{he2021masked} are introduced to predict a masked latent representation of patches. Similar ideas have been also explored in other recent works like BEiT~\cite{bao2021beit} and PeCo~\cite{dong2021peco} which adopt a discrete variational autoencoder (VAE) to generate discrete visual tokens from the original image. 
% Such kind of pretraining strategies worked well for self-supervised frameworks using vision transformer (ViT), achieving strong fine-tuning results on downstream application tasks like image classification, semantic segmentation etc. 
Motivated by these works, we expand this generative learning framework to tackle text recognition and document enhancement tasks.     
%~\cite{dong2021peco, bao2021beit}.

\noindent
\textbf{Text Recognition.}
Ample research in text recognition has been conducted, resulting in handwritten (HTR)~\cite{sonkusare2016survey, memon2020handwritten} and scene-text (STR)~\cite{shi2016end, long2021scene, chen2021text} recognition pipelines. Most common approaches that tackle text recognition are using supervised methodologies that employ an encoder-decoder mechanism~\cite{cheng2017focusing, shi2016end, shi2016robust, litman2020scatter, kang2020pay}
% ~\cite{cheng2017focusing, shi2016end, shi2016robust, sueiras2018offline, fogel2020scrabblegan, litman2020scatter, yousef2020origaminet, kang2020pay}
based on a Connectionist Temporal Classification (CTC)~\cite{graves2006connectionist} network or an Attention-based~\cite{cheng2017focusing, shi2016robust} decoder. 
% network to represent the input with a latent representation to later be used as input into a module~\cite{cheng2017focusing, shi2016end, shi2016robust, sueiras2018offline, fogel2020scrabblegan, litman2020scatter, yousef2020origaminet, kang2020pay} based on a Connectionist Temporal Classification (CTC)~\cite{graves2006connectionist} or Attention-based~\cite{cheng2017focusing, shi2016robust} decoder. 
% Other approaches often extract task-dependent features and directly use them in a downstream task such as retrieval~\cite{mafla2021real, wang2021scene}, classification\cite{mafla2020fine, mafla2021multi, bai2018integrating} and Visual Question Answering (VQA)\cite{biten2019scene, singh2019towards} among others. 
% A recent work~\cite{mou2020plugnet} tackled the problem of scene text recognition under degraded scenario using a super-resolution unit to enhance the degraded text at the feature-level during training. 
Recently, approaches that focus on semi-supervised and self-supervised learning have been explored~\cite{souibgui2021few} with domain adaptation techniques on STR~\cite{kang2020unsupervised} and HTR~\cite{zhang2019sequence}.
% Domain adaptation techniques are employed by~\cite{kang2020unsupervised} and~\cite{zhang2019sequence} to recognize handwritten and scene-text respectively. 
% Both approaches rely on closing the gap among datasets that are labeled with those unlabeled ones. 
Under the unsupervised paradigm,~\cite{gupta2018learning} formulate text recognition as a task to align the conditional distribution of strings predicted with lexically correct strings sampled from a text database. Closely related to our work,~\cite{aberdam2021sequence} proposes a self-supervised sequence-to-sequence model that separates consecutive text features to be later used in a contrastive loss similar to~\cite{chen2020simple}. Analogously,~\cite{zhang2022context} and ~\cite{liu2022perceiving} improve the features obtained from a contrastive loss by concatenating characters and by perceiving spatial strokes respectively. Nevertheless, these methods require large batches, and rely on a sequential definition of features that can produce misaligned characters or n-grams contained in different words.

% To the best of our knowledge, our model is the first pipeline that employs masking in an auto-encoder manner, which exploits the bidirectional attention of transformers to produce richer visual representations. Finally, our method yields state-of-the-art results in handwritten and scene-text recognition... or more tasks?

\begin{figure*}[t!]
\centering
\includegraphics[width=\linewidth, height=7cm]{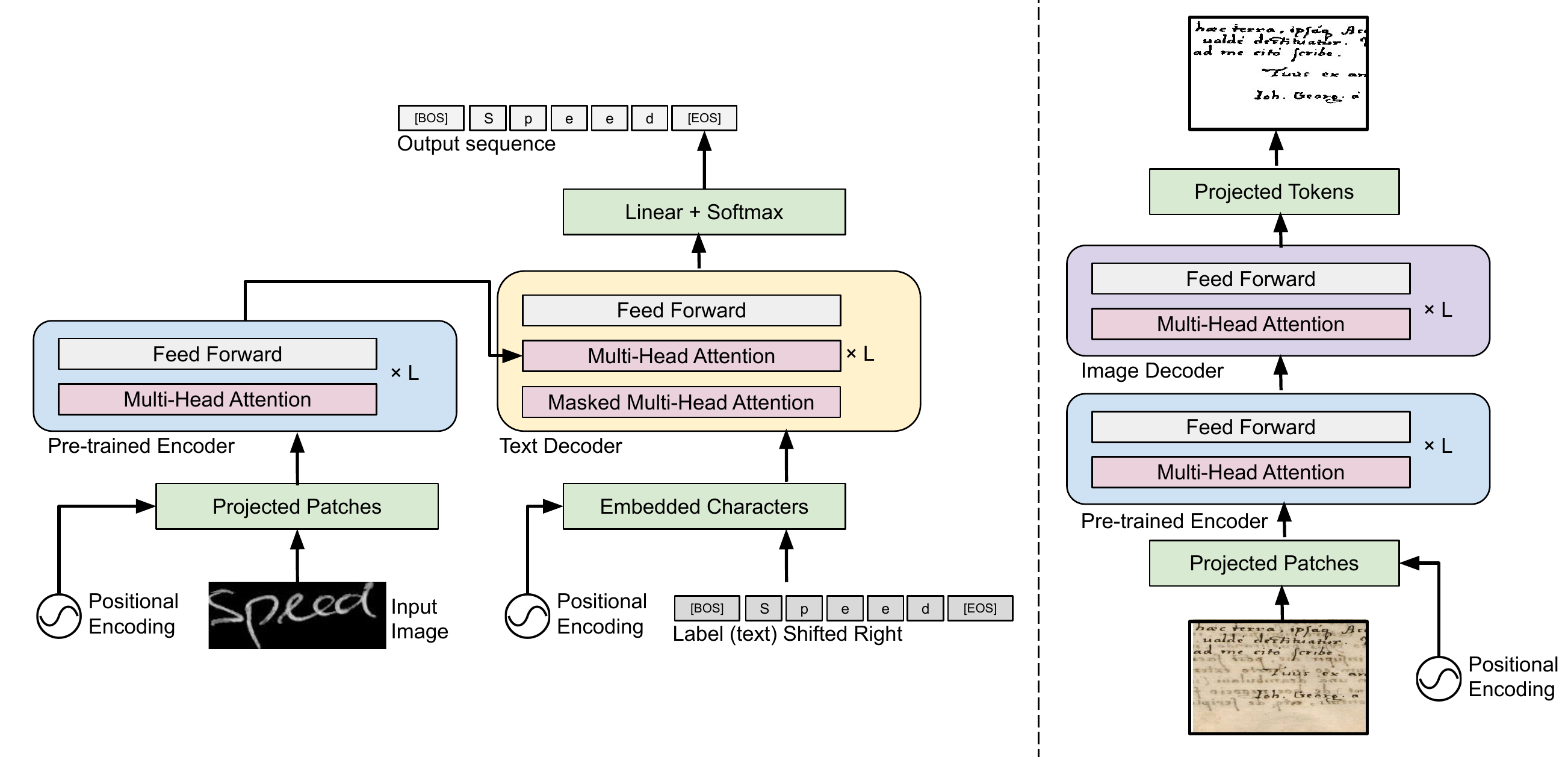}
\caption{\textbf{Fine-tuning pipeline}. We start from a pretrained encoder as initial weights to solve a specific downstream task. Explicit decoders are used for text recognition (left) and document image enhancement (right).}
\label{fig:method-ocr}
% \vspace{-.3cm}
\end{figure*}

\noindent
\textbf{Document Image Enhancement.}
Many approaches have been proposed to address the enhancement of documents (both handwritten and machine-printed) which suffer several kinds of artefacts/defects such as bleed-through, show-through, faint characters, contrast variations and so on. 
% This work proposes to tackle two important document enhancement tasks: binarization and deblurring. 
Adaptive thresholding based on sliding-window operations by~\cite{sauvola2000adaptive} formulated a strong handcrafted baseline for binarization task.
% remained a strong handcrafted baseline on this task for years. 
% The rise of deep learning brought newer approaches where
The work from~\cite{calvo2019selectional, kang2021complex} map images from the degraded domain to the enhanced one using end-to-end CNN-based autoencoders. Other techniques~\cite{zhao2019document, souibgui2020gan, souibgui2020conditional, jemni2022enhance} used conditional-Generative Adversarial Network (c-GAN) based approaches to design a generator which produces the enhanced version of the document while the discriminator assesses the quality of binarization. Lately, an end-to-end ViT autoencoder was proposed in~\cite{souibgui2022docentr} to capture high-level global features using self-attention for binarizing degraded documents. Regarding document deblurring, a benchmark was formulated by~\cite{hradivs2015convolutional} where a CNN were trained to reconstruct enhanced high-quality images from blurry inputs that consist of a combination of camera-driven motion blurred and de-focused images of text documents. Lately, \cite{souibgui2020gan} improved the baseline performance using the similar c-GAN based approach aimed in a binarization task.

%% file: tex/method.tex
\begin{figure}[t!]
\centering
\includegraphics[width=\linewidth]{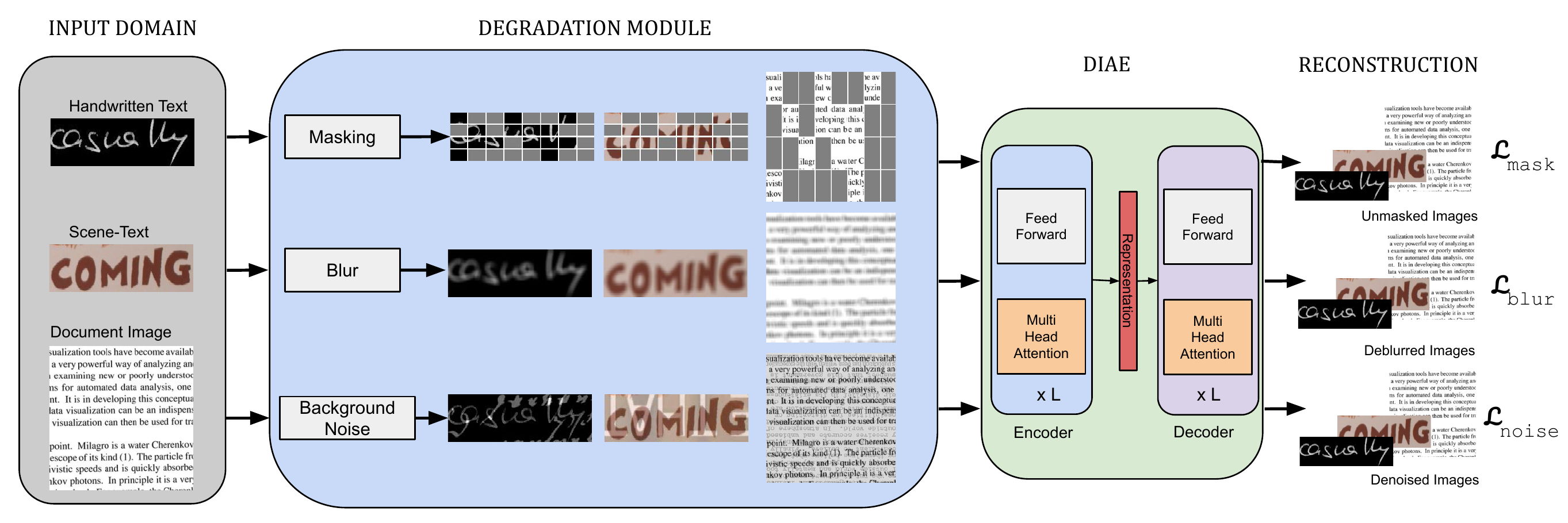}
\caption{\textbf{Pre-training pipeline.} 
% We perform degradation of an input image coming from a specific domain. 
Text-DIAE aims to learn degradation invariant representations. These are later used to reconstruct the input image with a specific learning objective for each degradation type.}
\label{fig:main_model}
% \vspace{-.4cm}
\end{figure}

In this section, we present our proposed method for text image recognition and enhancement by describing its building blocks. Our approach uses two steps: a pre-training stage to learn useful representations from unlabeled data, and a supervised fine-tuning phase for the desired downstream task.

%We begin by formulating the problem, then we describe the used architectures for each task.
%In what follows, We  formulate  the process and present the  

% Unsupervised learning exploits information
% redundancy and complementarity in the image data by learning to reconstruct local content along with contextual features. 
% learning to reconstruct local content by integrating it with
% context

% \begin{figure}
% \centering
% \includegraphics[width=\linewidth]{images/pretraining.png}
% \caption{3 pretext tasks.}
% \label{fig:method-ocr}
% \end{figure}

% In this section we introduce the studied scenarios and methods employed.  Our self-supervised autoencoder 
\subsection{Pre-Training Module}\label{seq:pretraining}
% firstly, we formulate  the pre-traning process, then we present the two different downstream tasks: Text recognition and document image enhancement. 

% \textbf{Pre-training}
%  As it can be seen, in each iteration,  an image  is fed to the degradation module. In this stage, three different degradation are inserted to the image.
The overall pre-training pipeline of Text-DIAE is shown in Fig.~\ref{fig:main_model}.
% We hypothesize that there exists a latent subspace which is invariant to changes in $Deg$, i.e. it is degradation invariant.  
For each task, given an unlabeled image $I$ (eg. a cropped handwritten text, cropped scene text or a scanned document image)%belonging to a specific task, 
, we use a function $\phi$ to map $I$ to a degraded form. The function $\phi$ takes as parameters the original image $I$ and the degradation type $\mathcal{T} \in \{mask, blur, noise\}$ where  
% and produce one of the 3 possible degraded images as shown in the degradation module of Figure~\ref{fig:teaser}. 
we denote a degraded image by $I_d = \phi(I, \mathcal{T})$. 
  
% The ability of our proposed DIAE to reconstruct the original input when exposed to different degraded image transformations $Deg$ are proposed as the pre-text tasks.   

% The degraded images are used to learn a restoration model that reconstructs $I_d$ to $I$.

Our model is composed of an encoder $\mathcal{E}$ and a decoder $\mathcal{D}$ with learnable parameters $\theta_\mathcal{E}$, $\theta_\mathcal{D}$ respectively. The pre-training pipeline trains an encoder function $\mathcal{E}$ that maps the degraded image $I_d$ to a latent representation $z_\mathcal{T}$ in a multi task fashion (unmasking, deblurring and denoising) and then learning a decoder $\mathcal{D}$ to reconstruct the original image $I$ from the representation $z_\mathcal{T}$: 
% as shown in Eqn.~\ref{eq:pretraining}  .
\begin{equation}
\begin{split}
    z_\mathcal{T} &= \mathcal{E}(\phi(I, \mathcal{T}); \theta_E) \\
    I_r &= \mathcal{D}(z_\mathcal{T}; \theta_D)
\end{split}
\label{eq:pretraining}
\end{equation}
% \begin{equation}\label{eq:decoding}
%     I_r = D(z, Deg; \theta_D)
% \end{equation}
% The goal of the pre-training is to learn that can maps  an inputted text image to the best possible representation, for later processing tasks.  
% The key objective of pre-training is to learn the most significant parameters $\theta_E$ from the latent estimation $z$ via three proposed pre-text tasks and without any manually defined annotations (in a self-supervised way). 
The learned visual representations from the latent subspace should be invariant to the applied degradation $\mathcal{T}$. 
% and would contain useful information to reconstruct the original input $I$. 
% In what follows we discuss the key components of DIAE in a more fine-grained manner.\\
\input{tables/frozen_encoder_text}

\noindent
\textbf{Encoder.}
The encoder architecture consists of a vanilla ViT~\cite{dosovitskiy2020image} backbone. Given an input image $I_d$, it is first split into a set of $N$ patches, $I_d^p = \{I_d^{p_1}, I_d^{p_2},\\ ... , I_d^{p_N}\}$. Then, these patches are embedded with a trainable linear projection layer $E$. 
% where 
% to obtain their representative tokens 
% $I_d^p E = \{{I_p}^1 E, {I_p}^2 E, ... , {I_p}^N E\}$. 
Text-DIAE uses a distinct linear projection layer for every defined pre-text task. These tokens are later concatenated with their 2-D positional information embedded with $E_{pos}$ and fed to $L$ transformer blocks to map these tokens to the encoded latent representation $z_l$. These blocks are composed of $L$ layers of Multi-head Self-Attention (MSA) and a feedforward Multi-Layered Perceptron (MLP) as depicted in Figure~\ref{fig:main_model}. Each of these blocks are preceded by a LayerNorm (LN)~\cite{ba2016layer} and followed by a residual connection: 
% % The aforementioned statement is exemplified in Eqn.~\ref{eq:encoding}
\begin{equation}
\begin{split}
    z_0 &=  E(I_d^p) + E_{pos} \\
    {z^\prime}_l &= \text{MSA}(\text{LN}(z_{l-1})) + z_{l-1} \text{, \space} l = 1, \text{ \dots L} \\
    {z}_l &= \text{MLP}(\text{LN}({z^\prime}_l)) + {z^\prime}_l \text{, \space} l = 1, \text{ \dots L} \\
    z_\mathcal{T} &= LN(z_L) 
    % z_\mathcal{T} \in R^{N\times D}
\end{split}
\label{eq:encoding}
\end{equation}

\noindent
\textbf{Decoder.} The decoder composed of transformer blocks following the same structure and number of layers as the encoder. The decoder input is 
% the sequence of  outputted tokens from 
the output of encoder $z_\mathcal{T}$. 
% These tokens are propagated into the transformer decoder blocks and then  projected with a linear layer to the desired pixel values. 
The output of the decoder is a set of vectors $I_{r} = \{I_{r}^{p_1}, I_{r}^{p_2}, ... , I_{r}^{p_N}\}$ where each of which corresponds to a flattened patch in the predicted (reconstructed) image. Same as before, a distinct linear layer is used for each pre-text task.
\begin{equation}
\begin{split}
    {z^\prime}_l &= \text{MSA}(\text{LN}(z_{l-1})) + z_{l-1} \text{ , \space} l = 1, \text{ \dots L} 
\\
    {z}_l &= \text{MLP}(\text{LN}({z^\prime}_l)) + {z^\prime}_l \text{ , \space} l = 1, \text{ \dots L}
\\
  I_{r} &= \text{Linear}(z_L)
\end{split}
\label{eq:decoding}
\end{equation}

\subsection{Fine-Tuning}

Our fine tuning process is illustrated in Fig.~\ref{fig:method-ocr} where we perform two different downstream tasks; text recognition and document image enhancement.
% Our fine tuning  is illustrated in Fig.~\ref{fig:method-ocr}. in this stage, we perform two different downstream tasks, text recognition and document image enhancement. We detail each of them in what follows.
% We use the same encoder as described in the previous Encoder section.
% Describe the differences of the Decoder compared to pre-training
\input{tables/finetuning_text}

\noindent
\textbf{Text Recognition.} 
% The goal of text recognition is to transform a text image into a machine encoded form (sequence of characters), this is usually done by training a machine learning model in a supervised way. 
Text recognition aims to transform an image into the machine encoded form, i.e.,  sequence of characters.
Let $I$ be a cropped text image and $C_{}=\{c_{{}_1},c_{{}_2}, ..., c_{{}_N}\}$ its ground truth label which corresponds to a sequence of characters, where $N$ is the length of the text. The training is done by passing $I$ to an encoder function $\mathcal{E}$ to produce a latent representation $z$. Then, $z$ is later fed to a decoder function $\mathcal{D^{\prime}}$ to produce a sequence of characters $C_{p}=\{c_{p_1},c_{p_2}, ..., c_{p_N}\}$ that should match the ground truth label sequence. 
% In this way, we learn $\theta_E$ and $\theta_D$ as follows:
% \begin{equation}
%     [c_{p_1},c_{p_2}, ..., c_{p_N}] = D(z;\theta_D) \text{, \space } z = E(I;\theta_E) 
% \end{equation}

% In our formulation, since we are doing a fine tuning,  we use the same encoder as in the previous pre-training stage as seen in Fig.~\ref{fig:method-ocr}-Left. 
% We initialize the encoder with the pre-trained weights $\theta_E$ as a starting point for this learning process. For the decoder, we employ a sequential transformer decoder, same as \cite{vaswani2017attention}. 
We initialize the encoder with the pre-trained weights $\theta_\mathcal{E}$ while we employ a sequential transformer decoder~\cite{vaswani2017attention} as seen in Fig.~\ref{fig:method-ocr}-Left. The decoder is initialized randomly and composed of $L$ transformer blocks of MSA, MLP and Masked-MSA layers preceded by LN layers, and followed by a residual connection.
% as in the pre-training decoder, preceded by LN layers, and followed by a residual connection. 
The output of the decoder is a sequence of characters where at each time step $t$, the predicted character is formed by attending to the representation $z$ and previous character embeddings until $t-1$. 
% $\{c_{1},c_{2}, ..., c_{t-1}\}$.
% encoded by $E$ as detailed in Eqn~\ref{eq:encoding} and the GT characters $\{c_{1},c_{2}, ..., c_{t-1}\}$ followed by a linear projection and a Softmax activation function.
% \begin{equation}
%     z_t = \text{Masked-MSA}([c_{{}_1}E,c_{{}_2}E, ..., c_{{}_N}E] + E_{pos}) + 
% \end{equation}
% \begin{equation}
%     z_t = [c_{{}_1}E,c_{{}_2}E, ..., c_{{}_N}E] + E_{pos}
% \end{equation}
% \begin{equation}
%     z_t = [c_{{}_1}E,c_{{}_2}E, ..., c_{{}_N}E] + E_{pos}
% \end{equation}

\noindent
\textbf{Document Image Enhancement. } Document enhancement consists of mapping a degraded document into a clean form. 
% as an image to image translation task. 
Let $I_d$ be a degraded image and $I_c$ its clean version, then the goal is to learn an encoder function $\mathcal{E}$ that maps $I_d$ to a representation $z$ with the same way as in Eqn~\ref{eq:encoding}. $\mathcal{E}$ weights are initialized from the pre-training stage. The decoder $\mathcal{D}^{\prime\prime}$ generates the clean image $I_c$ from $z$ as in Eqn~\ref{eq:decoding}. 
% this process can be formalized as:
% \begin{equation}
%     I_c = D(z;\theta_D) \text{ ,where } z = E(I_d;\theta_E) 
% \end{equation}
% In fine-tuning, $\theta_\mathcal{E}$ are initialized from the pre-training stage.
% We propose a two-step training procedure for the downstream application tasks: text recognition and document image enhancement. 
% \subsection{Approach: Pre-training}
% \subsection{Approach: Fine-Tuning}
% \dots

\subsection{Learning Objectives}

Our model makes use of different sets of losses for each phase. During pre-training, we use three different losses. Each one is dedicated to a particular pre-text task: $\mathcal{L}_{mask}$, $\mathcal{L}_{blur}$ and $\mathcal{L}_{noise}$. Each of these losses is a mean squared error (MSE) between the reconstructed image $\mathrm{I}_{r}$ (from the masked, blurred or noisy image) and its ground-truth version $\mathrm{I}_{gt}$. 
% Each of these losses is a mean squared error (MSE) between the reconstructed image and the masked, blurred or noisy image.
% Means, if we note the recovered image from a specific degradation by $I_{r_{Deg}}$ and the GT image by $I_{gt}$  $\mathcal{L}_{mask} = 1/n\sum{(I_{r_{mask}},I_{gt})}^2$,
% $\mathcal{L}_{blur} = 1/n\sum{(I_{r_{blur}},I_{gt})}^2$ and $\mathcal{L}_{noise} = 1/n\sum{(I_{r_{noise}},I_{gt})}^2$.
Thus, the overall loss for our pre-training stage is:
\begin{equation}
    % \mathcal{L}_{pre-training} = \mathcal{L}_{mask} + \mathcal{L}_{blur} + \mathcal{L}_{noise}
    \resizebox{0.45\textwidth}{!}{$\mathcal{L}_{pt} =\lambda_{1}\mathcal{L}_{m}\left(\mathrm{I}_{r}, \mathrm{I}_{gt}\right)+\lambda_{2} \mathcal{L}_{b}\left(\mathrm{I}_{r}, \mathrm{I}_{gt}\right) 
    +\lambda_{3} \mathcal{L}_{n}\left(\mathrm{I}_{r}, \mathrm{I}_{gt}\right)$}
\end{equation}
During our experimentation, the best results were obtained with setting $\lambda_{1}=\lambda_{2}=\lambda_{3}=1$.

While fine-tuning on text recognition, we use a cross-entropy loss between the predicted sequence of characters $C_p$ and $C$. 
% Thus, the training loss is:
% \begin{equation}
%     \mathcal{L}_{recognition} = - 1/n \sum C_i \text{log}(C_{i_p}) + (1-C_i) \text{log}(1-C_{i_p})
% \end{equation}
For document image enhancement fine tuning, we used an MSE loss between the cleaned image $I_c$ and $I$. 
% Thus:
% \begin{equation}
%     \mathcal{L}_{enhancement} = 1/n\sum{(I_{r_{}},I_{gt})}^2
% \end{equation}
% \textbf{Pre-training}
% MSE for each

% \textbf{Fine-tuning}
% Cross Entropy for text and MSE for image docs

%% file: tables/frozen_encoder_text.tex
\setlength{\tabcolsep}{2pt}
\begin{table*}[t!]
\footnotesize
\begin{center}

\caption{\textbf{Representation quality.} 
We evaluate the encoder capability of learning visual representations. This scenario is analogous as the linear probing in self-supervised models. We train a decoder with labelled data on top of a frozen encoder pre-trained on the proposed degradation. The column \textit{Seen} refers to the number of samples in millions seen during pre-training. Word prediction in terms of Accuracy (Acc) and single edit distance (ED1) in handwritten and text recognition. } 
\label{table:results_frozen_encoder}
\resizebox{\textwidth}{!}{
\begin{tabular}{lcccccccccccccc}
\hline
\multirow{3}{*}{Method} & \multirow{3}{*}{Encoder} & \multirow{3}{*}{Decoder} & \multicolumn{6}{c}{Handwritten Text} & \multicolumn{6}{c}{Scene-Text} \\
 &  &  & \multicolumn{3}{c}{IAM}  &\multicolumn{3}{c}{CVL} & \multicolumn{3}{c}{IIIT5K} & \multicolumn{3}{c}{IC13} \\ %\hline
 & & & Acc & ED1 & Seen &Acc & ED1 & Seen & Acc & ED1 & Seen & Acc & ED1& Seen  \\ \hline
simCLR~\cite{chen2020simple} & \multirow{3}{*}{CNN} & \multirow{4}{*}{CTC} & 4.0 & 16.0& 205.8 &1.8 &11.1& 205.8 & 0.3& 3.1&409.6 & 0.3& 5.0&409.6 \\
seqCLR~\cite{aberdam2021sequence} &  &  & 39.7& 63.3 &205.8 & 66.7& 77.0& 205.8 & 35.7& 62.0 &409.6& 43.5 & 67.9&409.6\\ 
PerSec~\cite{liu2022perceiving} &  &  & --&-- &-- & --& --& -- &37.9& -- &--& 46.4 & -- & --\\ 
PerSec~\cite{liu2022perceiving} & ViT &  & --&-- &-- & --& --& -- &38.4& -- &--& 46.7 & -- & --\\ \hline
simCLR~\cite{chen2020simple} & \multirow{3}{*}{CNN} & \multirow{4}{*}{Attn.} &  16.0&21.2& 205.8 & 26.7& 30.6 &205.8 & 2.4& 3.6&409.6 & 3.1& 4.9&409.6 \\
seqCLR~\cite{aberdam2021sequence} &  &  & 51.9& 65.0 &205.8& 74.5& 77.1 &205.8& 49.2& 68.6&409.6 & 59.3& 77.1&409.6 \\
PerSec~\cite{liu2022perceiving} &  &  & --&-- &-- & --& --& -- &50.7& -- &--& 61.1 & -- & --\\ 
PerSec~\cite{liu2022perceiving} & ViT &  & --&-- &-- & --& --& -- &52.3& -- &--& 62.3 & -- & --\\ \hline
\textbf{Ours} & ViT & Transf. & \textbf{71.0} & \textbf{82.1}& \textbf{4.7} &\textbf{78.1} &\textbf{81.5} & \textbf{1.2} & \textbf{77.1}  &\textbf{87.8} & \textbf{9.1} &\textbf{92.6}&\textbf{95.6}& \textbf{18.2}  \\ \hline

\end{tabular}}
\end{center}
% \vspace{-0.5cm}
\end{table*}
\setlength{\tabcolsep}{1.4pt}

%% file: tables/finetuning_text.tex
\setlength{\tabcolsep}{2pt}
\begin{SCtable*}
\footnotesize
% \scriptsize
% \begin{center}
\caption{\textbf{Semi-supervised results.} Accuracy obtained by fine-tuning a pre-trained model with varying percentages of the labeled dataset. Under this setting, we back-propagate the gradients through the specific decoder and the pre-trained encoder.}
\label{table:results_fine_tuning}
% \resizebox{\textwidth}{!}{
\begin{tabular}{lcccccccccc}
\hline\noalign{\smallskip}
\multirow{3}{*}{ Method } & \multirow{3}{*}{ Encoder } & \multirow{3}{*}{ Decoder } & \multicolumn{6}{c}{Handwritten Text} & \multicolumn{2}{c}{Scene-Text} \\
& & & \multicolumn{3}{c}{IAM} & \multicolumn{3}{c}{CVL} & IIIT5K & IC13 \\
& & & 5\% & 10\% & 100\% & 5\% & 10\% & 100\% & 100\% & 100\% \\
\noalign{\smallskip}
\hline
\noalign{\smallskip}
Supervised~\cite{aberdam2021sequence} & \multirow{4}{*}{CNN} & \multirow{5}{*}{ CTC } & 21.4 & 33.6 & 75.2 &48.7 & 63.6  &75.6 & 76.1 & 84.3 \\
simCLR~\cite{chen2020simple} & & & 15.4 & 21.8 & 65.0 &52.1 & 62.0  &74.1  & 69.1 & 79.4 \\
seqCLR~\cite{aberdam2021sequence} & & & 31.2 & 44.9 & 76.7 & 66.0 & 71.0 & 77.0 & 80.9 & 86.3 \\ 
 {PerSec \cite{liu2022perceiving}} & & & -- & -- & 77.9 & -- & -- & 78.1 & 82.2 & 87.9 \\ 
 {PerSec \cite{liu2022perceiving}} & ViT& & -- & -- & 78.0 & -- & -- & 78.8 & 83.7 & 89.7 \\ 
\noalign{\smallskip}
\hline
\noalign{\smallskip}
Supervised~\cite{aberdam2021sequence} & \multirow{4}{*}{ CNN } & \multirow{5}{*}{ Attn. } & 25.7 & 42.5 & 77.8 &  64.0 & 72.1 & 77.2& 83.8
& 88.1 \\
simCLR~\cite{chen2020simple} & & & 22.7 & 32.2 & 70.7 & 59.0& 65.6&  75.7 & 77.8 & 84.9 \\
seqCLR~\cite{aberdam2021sequence} & & & 40.3 & 52.3 & 79.9 & \textbf{73.1} & \textbf{74.8}& 77.8 & 82.9 & 87.9 \\
 {PerSec \cite{liu2022perceiving}} & & & -- & -- & 80.8& -- & -- & 80.2 & 84.2& 88.9 \\ 
 {PerSec \cite{liu2022perceiving}} & ViT& & -- & -- & \textbf{81.8} & -- & -- & 80.8 & 85.2 & 89.2 \\ 
\noalign{\smallskip}
\hline
\noalign{\smallskip}
Supervised (Ours) & \multirow{2}{*}{ ViT } & \multirow{2}{*}{ Transf.} & 22.8 & 25.3 & 71.7 &17.9 & 19.8 &71.9 & 75.7& 91.9\\
\textbf{Ours} & & & \textbf{49.6} & \textbf{58.7} & {80.0} & 47.9 &68.5 &\textbf{87.3} &\textbf{86.1} & \textbf{92.0}\\
% Ours-Large & & & & & & & & & & \\
\hline
\end{tabular}
% }
% \end{center}
% \vspace{-0.5cm}
\end{SCtable*}
\setlength{\tabcolsep}{2pt}

%% file: tex/experiments.tex
In this section we describe the studied scenarios and experiments performed for text recognition and document enhancement respectively. We ask the reader to refer to the supplementary material for specific implementation details.

\subsection{Text Recognition}
\noindent
\textbf{Evaluating Representations.} In order to evaluate the quality of the learned representations, and extending commonly used linear-probing settings~\cite{zhang2016colorful}, we employ a similar approach as introduced by~\cite{aberdam2021sequence}. As a first step, the encoder is pre-trained with unlabeled data as described in Section \ref{seq:pretraining}. After that, the encoder's weights are frozen and a new decoder is trained on top of it with all the labeled data. The decoder, as we detailed above, generates the predicted characters in a time-step manner. Since the encoder remains frozen, this scenario is a good proxy that represents the expressivity of the learned visual representations. To this end, Table~\ref{table:results_frozen_encoder} shows the results of our proposed approach. We compare among self-supervised methods specifically designed for the text recognition task.

\begin{figure}[t!]
\centering
\includegraphics[width=\linewidth]{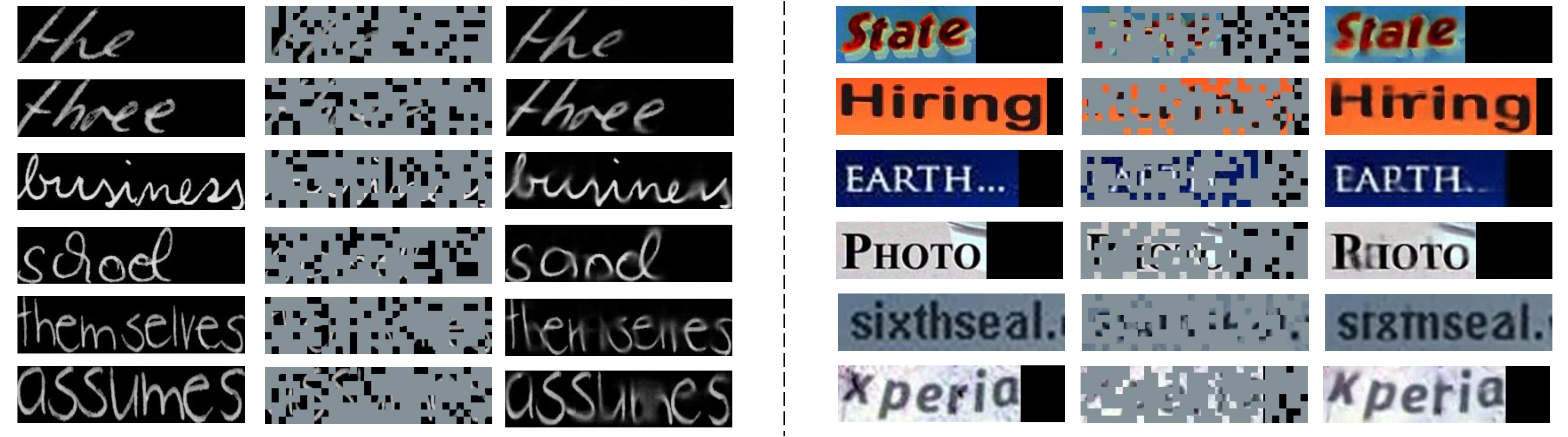}
\caption{\textbf{Qualitative results of pre-training samples.} The left refers to handwritten text, while scene-text is depicted on the right. On each scenario, from left to right, the original, masked and reconstructed images are depicted.}
\label{fig:qualitative_ocr}
% \vspace{-.3cm}
\end{figure}

\input{tables/two_tables}
\noindent
\textbf{Better performance.}
% achieved by Text-DIAE.}
As it can be seen from Table~\ref{table:results_frozen_encoder}, the seqCLR method presented by~\cite{aberdam2021sequence} improves significantly a self-supervised baseline inspired by SimCLR~\cite{chen2020simple}. In the recently released approach PerSec by~\cite{liu2022perceiving}, they slightly improve over the seqCLR.  
% The improvements presented by SeqCLR~\cite{aberdam2021sequence} come from incorporating a contrastive loss on top of the sequential features. 
% However, 
It is evident that our Text-DIAE model \textit{greatly} outperforms all the aforementioned state-of-the-art approaches regarding the representation quality obtained, both in handwritten and scene-text. The improvements in term of the accuracy in a handwritten text dataset, IAM, is close to \textbf{+20 points}. Moreover, a bigger improvement gap is obtained when evaluating scene-text. An average gain of \textbf{+30 points} is accomplished in IIIT5K and ICDAR13, proving the generalization of our method to different domains.
In our model, the great expressivity of features achieved by the encoder is mainly due to two factors. 
Firstly, by masking image patches, the encoder learns a strong unigram character distribution (refer to Figure~\ref{fig:qualitative_ocr}), which is not leveraged in previous methods. Secondly, by  distorting and recovering  the image, we make the model learn richer representations to detect and recover the text into a clean and readable state. Thus, the model is learning the most valuable features that lead to the best recognition performance.

% employing a transformer as encoder, the self-attended embedding in the masking task allows the  

\noindent
\textbf{Faster convergence.} One of the most important outcomes by employing our method, is that a \textbf{paramount} improvement in convergence is achieved during pre-training.  Table~\ref{table:results_frozen_encoder} shows this effect under the column labeled as ``Seen''.  It  depicts the total number of seen samples that each model requires during the pre-training stage. It is worth highlighting that during pre-training the encoder of our model requires \textbf{~43} and \textbf{~166} times lesser data in IAM and CVL respectively when compared to the seqCLR and simCLR. In scene-text, our model employs only $18.2$M samples to yield powerful representations compared to the $409$M samples required by previous self-supervised approaches. 
% Shall we mention about early stopping? Just to say that probably better representations can be reached.
% \vspace{-1.cm}

\noindent
\textbf{Fine-Tuning.} In this stage, we evaluate our model considering a semi-supervised setting where the obtained results are depicted in Table~\ref{table:results_fine_tuning}. Here we use the self-supervised pre-trained encoder as a backbone  and train a transformer-based decoder from scratch that predicts the characters in a sequential manner, as illustrated in Fig.~\ref{fig:method-ocr}-Left. In this scenario, the gradients are back-propagated not only to the decoder but also to the encoder. Following the previous work~\cite{aberdam2021sequence}, we use 5\% and 10\% of the labeled dataset by randomly selecting the training samples. As suggested in~\cite{chen2020simple} we perform fine-tuning on all the labeled dataset. In order to compare with~\cite{aberdam2021sequence} and since scene-text dataset is synthetic, we evaluate with the complete labeled dataset.

\input{tables/ablation_htr_objective}
\noindent
\textbf{Higher performance in fine-tuning settings.} Our model exploits data in a more efficient manner than previous self-supervised methods in fine-tuning setting. We infer that the set of degradations proposed yields rich signals, helping the encoder to adapt to the downstream task more efficiently. Our model achieves state-of-the-art in all scenarios when all the labeled datasets are used except in IAM where the PerSec is slightly better.
Under semi-supervised settings, our model performs better at the IAM dataset when employing 5\% and 10\% of the labels than simCLR and seqCLR. Since CVL contains substantially fewer data samples than IAM, SeqCLR still outperforms our approach in the CVL dataset. However, while employing the full labels of CVL, Text-DIAE outperforms all the methods by a large margin. 
% This suggests that our approach may require more data points to leverage this advantages, however once a specific number of data samples point are seen, a more optimal solution is found.

\noindent
\textbf{More efficient than a supervised baseline.}
From table~\ref{table:results_fine_tuning}, we can also notice the superiority of pre-training our architecture  compared to a fully supervised model starting from scratch. 
%It is paramount to note the advantages that pre-training with the proposed degradation yield when comparing to such supervised model.
This suggest that the self-supervised pre-training of such transformer-based architectures is essential to obtain better results, and beneficial  especially in small labeled datasets scenarios, since the unlabeled data is generally easier to obtain for a self-supervised pre-training. 

\noindent
\textbf{The effect of fine-tuning after pre-training.} By proposing the degradation invariant optimization at pre-training, our model achieves a significant gain in recognition on handwritten text datasets. An average of 10 points of accuracy are gained after fine-tuning (refer to Table~\ref{table:results_frozen_encoder} and~\ref{table:results_fine_tuning}). 
Finally, it is important to note that our model reaches state-of-the-art in the handwritten text recognition task, even comparing to specifically designed supervised approaches. The results on the IAM dataset are shown in Table~\ref{table:results_fine_tuning_iam}, which measures the performance of a model in terms of word and character error rate, WER and CER respectively. 

% \vspace{-0.5cm}
\noindent
\textbf{Ablation Studies.}
The results of experimentation regarding the effect of each degradation as pretext task at pre-training is given in Table~\ref{tab:ablation_htr_objectives}. Firstly, among the three proposed degradations, masking is the most crucial to be applied in both tasks, handwritten and scene text recognition. 
% We believe that this is due to the inherent task of masking. 
When an input word is masked, and in order to properly reconstruct it, the model has to learn a character level distribution.
% of a given language. 
This by itself provides with a strong prior compared to denoising or deblurring an image. Additionally, adding blur in scene-text imagery improves the representations learned by the model shown by the results. Lastly, adding noise does not result in an improvement in text recognition tasks. However, as it is shown in the next section, the combination of the $3$ degradation produce a richer encoder in document enhancement. Therefore, we can safely assume that each degradation has a task-dependent impact on the representations learned depending on the similarity of them when compared to the final downstream task and input data distribution.

\input{tables/two_tables_enhance}
\noindent
\textbf{Qualitative Results.}
In Figure~\ref{fig:qualitative_ocr} we show the reconstructed images at pre-training stage for handwritten and scene-text samples. It is important to note the complexity of the reconstruction task even for humans. Even though high masking percentages are employed (75\%), our model learns to properly adapt to handwritten styles and fonts found in scene-text. As can be appreciated, although sometimes our model's reconstruction does not match with the ground truth images, it can still reconstruct the most probable and plausible English words (e.g. see ``school'' vs ``sand'' in 4th row in handwritten examples). Another interesting outcome is also noticed for scene-text example where ``xperia'' is reconstructed correctly while the last character ``a'' is selected from another font, demonstrating the model's capability.
Minor reconstruction errors are found such as that the model eventually learns to overcome at fine-tuning stage. 

\input{tables/dibco_table}
\input{tables/ablation_enhance_objectives}
\begin{figure*}[t]
\centering
\setlength{\tabcolsep}{0.31em}
\begin{tabular}{c c c c}
     \toprule
     \scriptsize{Original Input} & \scriptsize{DocEnTr~\cite{souibgui2022docentr}} & \scriptsize{\textbf{Ours}} & \scriptsize{Ground Truth} \\
     \midrule
     \includegraphics[width=0.23\linewidth]{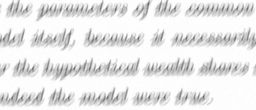} &  \includegraphics[width=0.23\linewidth]{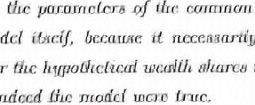} & \includegraphics[width=0.23\linewidth]{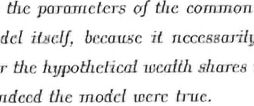} &
     \includegraphics[width=0.23\linewidth]{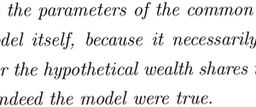} \\
     \midrule 
     \includegraphics[width=0.23\linewidth]{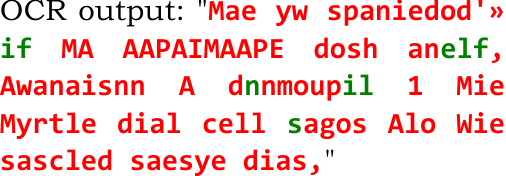} &  \includegraphics[width=0.23\linewidth]{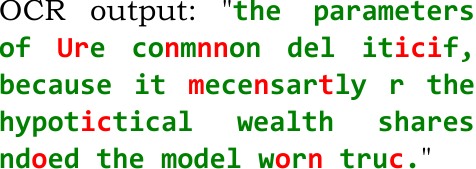} & \includegraphics[width=0.23\linewidth]{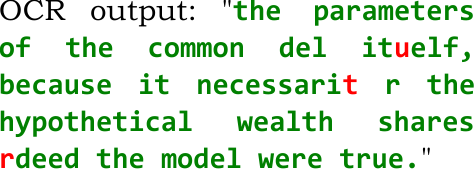} &
     \includegraphics[width=0.23\linewidth]{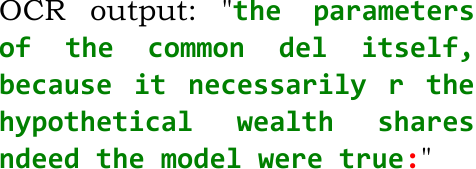} \\
     \midrule
     \scriptsize{CER: 78.86} &  \scriptsize{CER: 18.51} & \scriptsize{CER: \textbf{8.94}} & \scriptsize{CER: 4.88} \\
     \bottomrule
     \end{tabular}
\caption{\textbf{Qualitative results of deblurred samples.} The document image on the left refers to the originally captured blurred image, followed by the ground-truth, and the deblurred results from the DocEnTr and our Text-DIAE model towards right. The correctly predicted OCR  output is shown in "Green" font while the inaccurate ones are depicted in "Red" and recognition performance in terms of CER.}
\label{fig:ocr_deblur}
% \vspace{-.5cm}
\end{figure*}

\noindent

\subsection{Document Image Enhancement}
% \textcolor{blue}{After validating our method on the HTR and STR tasks, we test in in this section on two more downstream applications: Binarization and deblurring.}

\noindent
\textbf{Performance Analysis on Binarization.}
% After pre-training, we explore the credibility of the learned representations
% %during the pre-training procedure for two 
% for two downstream applications: Binarization and deblurring. 
As shown in Table~\ref{tab:dibco_sota}, the Text-DIAE outperforms the previous state-of-the-art approaches on majority of the standard metrics for document binarization task. Specifically, the quantitative comparison of results demonstrate that Text-DIAE achieves an optimal gain in PSNR, FM, $F_{ps}$ and DRD performance surpassing the all previous arts. The largest performance improvement is obtained over the H-DIBCO 2012 while the least performance gain is obtained in the H-DIBCO 2018. 
% which implies that it is the most challenging one.  
% The results clearly demonstrate how the self-supervised Text-DIAE outperforms other techniques based on image thresholding~\cite{sauvola2000adaptive},  CNNs~\cite{zhao2019document}, conditional GANs~\cite{kang2021complex} and most recent ViT autoencoders~\cite{souibgui2022docentr}. 
One of the major concerns which degraded historical documents face is the show-through effect, which appears when ink impressions from one side of the document start appearing on the other side, making it almost impossible to read as shown in Appendix. 
% Figure~\ref{fig:binarization_qualitative_2}.
% In Fig.~\ref{fig:binarization_qualitative_2} we also show a qualitative comparison of an example case study taken from the DIBCO 2017 dataset. 
The enhanced Text-DIAE output 
% in Figure~\ref{fig:binarization_qualitative_2} 
illustrates that it not only resolves the show-through but also sharpens and smoothens the edges of the foreground text approximately to the ground-truth level. 

\noindent
\textbf{Performance Analysis on Deblurring.}
In Table~\ref{tab:results_deblurring} we show a quantitative comparison and superiority of Text-DIAE over supervised techniques ~\cite{hradivs2015convolutional,wang2018high, souibgui2020gan, souibgui2022docentr} on the document deblurring benchmark. A substantial gain in PSNR by \textbf{+2 points} on a \textbf{logarithmic} scale is obtained over DocEnTr~\cite{souibgui2022docentr}, which signifies the greater quality of deblurred images generated by Text-DIAE. There are two different kinds of blurring which appear in documents: motion blur owing to the sudden rapid camera movement and out-of-focus blur which emerges when light fails to converge in the image. In Fig.~\ref{fig:ocr_deblur}, we show an interesting qualitative case study of a motion blurred document image. We assess the performance of deblurring by running the Tesseract-OCR engine~\cite{smith2007overview} over the blurred, ground-truth, DocEnTr prediction and the Text-DIAE output. Qualitative results show that Text-DIAE significantly decreases the CER, showing vast improvement in OCR performance as depicted in green font. 
% Also, Text-DIAE achieves a very decent recognition performance on the deblurred sample produced by the Text-DIAE as depicted in green font.     
% \input{tables/deblurring_table}

\noindent
\textbf{Ablation Studies.} We also showcase an interesting ablation on the task of document image binarization for the challenging DIBCO 2018 benchmark. 
From Table~\ref{tab:ablation_enhance_objectives}, we infer that any pre-training task is beneficial while the denoising task is the most crucial to be applied when each pre-text task is applied separately. The aforementioned result can be attributed to the fact that denoising is much closer to the downstream binarization task. Also, it demonstrates that Text-DIAE performs the best for document enhancement tasks when the model learns all the possible degradation (masking, blurring and adding noise) together.
% in the representation space.    
% \\ \textbf{Denoising task is more beneficial for document image binarization:} In the quantitative comparison, we observe that the denoising pre-training objective alone is more effective than masking and deblurring ones.  
% \\ \textbf{Combining all degradations learn the best features:} The results also demonstrate that DIAE performs the best for binarizing document images  with learning representations from all kind of degradations(mask, blur and noise) as pretext tasks. 
% \input{tables/ablation_loss_enhance}

% \begin{figure}[h]
% \centering
% % \includegraphics[width=\linewidth,scale=0.25]{images/Sample_Tesseract.pdf}
% \includegraphics[width=0.7\textwidth, scale=0.2]{images/Sample_Tesseract.pdf}
% \caption{Fine tuning with a pretrained encoder for different downstream tasks. Left: Text Recognition. Right: Document image enhancement. }
% \label{fig:tesseract-ocr-pred}
% \end{figure}

% \subsubsection{Improving OCR}

%% file: tables/two_tables.tex
\begin{table}[t!]
% \begin{minipage}{.35\textwidth}

\setlength{\tabcolsep}{2pt}
% \scriptsize
% \tiny
\begin{center}
\caption{\textbf{SOTA results.} 
%Word and character error rates (WER and CER) of the proposed method compared to 
Quantitative evaluation with state-of-the-art methods on the IAM word level dataset.}
\label{table:results_fine_tuning_iam}
\begin{tabular}{lccc}
\hline\noalign{\smallskip}
Method & CER$\downarrow$ & WER$\downarrow$ & Avg. \\
\noalign{\smallskip}
\hline
\noalign{\smallskip}
Bluche et al. \cite{bluche2015deep}   & \textbf{7.3}& 24.7& 16.00 \\
Bluche et al. \cite{bluche2016joint}  & 7.9& 24.6 & 16.25 \\
Sueiras et al.  \cite{sueiras2018offline} & 8.8& 23.8 & 16.30 \\
ScrabbleGAN \cite{fogel2020scrabblegan}  & -& 23.6 & - \\
SSDAN \cite{zhang2019sequence}  & 8.5 & 22.2& 15.35 \\
SeqCLR \cite{aberdam2021sequence} & 9.5& 20.1  & 14.80 \\
PerSec \cite{liu2022perceiving} & -& 18.2 & - \\
\textbf{Ours}  & 9.3 & \textbf{20.0}& \textbf{14.65} \\
% Ours-Large & & & \\
\hline
\end{tabular}
\end{center}
% \vspace{-.6cm}
\end{table}

% \begin{table}
% \setlength{\tabcolsep}{2pt}
% % \tiny
% \centering
% \caption{\textbf{Ablations of the pre-training objectives.} Results in handwritten and scene-text recognition obtained by each pretext task. The performance is measured in terms of Word and Character error rates (WER and CER).}
% \label{tab:ablation_htr_objectives}
% \resizebox{\columnwidth}{!}{
% \begin{tabular}{ccccccccc}
% \hline\noalign{\smallskip}
% \multirow{2}{*}{$\mathcal{L}_{mask}$} & \multirow{2}{*}{$\mathcal{L}_{blur}$} & \multirow{2}{*}{$\mathcal{L}_{noise}$} & \multicolumn{3}{c}{IAM}& \multicolumn{3}{c}{IC13}\\
% & & & CER$\downarrow$ & WER$\downarrow$ & Avg. &CER$\downarrow$ & WER$\downarrow$ & Avg. \\
% \noalign{\smallskip}
% \hline
% \noalign{\smallskip}
% %\xmark &\xmark&\xmark& 11.8 & 31.3 & 21.55 \\
% \cmark &\xmark&\xmark& \textbf{9.3}	&\textbf{20.0}	&\textbf{14.65} & 4.5 & 8.0 & 6.25 \\
% \cmark &\cmark&\xmark& 12.3	&24.8	&18.5 & \textbf{4.2}	& \textbf{8.0} &	\textbf{6.10} \\
% \cmark &\xmark&\cmark& 11.1	&23.3	&17.2& 4.8&	8.6	&6.70 \\
% \cmark &\cmark&\cmark& 11.4	&23.8 &	17.6 & 5.1&	9.3	&7.20 \\
% \hline
% \end{tabular}}
% \setlength{\tabcolsep}{1.4pt}
% \vspace{-.5cm}
% \end{table}

%% file: tables/ablation_htr_objective.tex
\begin{table}[t!]
\setlength{\tabcolsep}{2pt}
% \tiny
\centering
\caption{\textbf{Ablations of the pre-training objectives.} Results in handwritten and scene-text recognition obtained by each pretext task. The performance is measured in terms of Word and Character error rates (WER and CER).}
\label{tab:ablation_htr_objectives}
\resizebox{\columnwidth}{!}{
\begin{tabular}{ccccccccc}
\hline\noalign{\smallskip}
\multirow{2}{*}{$\mathcal{L}_{mask}$} & \multirow{2}{*}{$\mathcal{L}_{blur}$} & \multirow{2}{*}{$\mathcal{L}_{noise}$} & \multicolumn{3}{c}{IAM}& \multicolumn{3}{c}{IC13}\\
& & & CER$\downarrow$ & WER$\downarrow$ & Avg. &CER$\downarrow$ & WER$\downarrow$ & Avg. \\
\noalign{\smallskip}
\hline
\noalign{\smallskip}
%\xmark &\xmark&\xmark& 11.8 & 31.3 & 21.55 \\
\cmark &\xmark&\xmark& \textbf{9.3}	&\textbf{20.0}	&\textbf{14.65} & 4.5 & 8.0 & 6.25 \\
\cmark &\cmark&\xmark& 12.3	&24.8	&18.5 & \textbf{4.2}	& \textbf{8.0} &	\textbf{6.10} \\
\cmark &\xmark&\cmark& 11.1	&23.3	&17.2& 4.8&	8.6	&6.70 \\
\cmark &\cmark&\cmark& 11.4	&23.8 &	17.6 & 5.1&	9.3	&7.20 \\
\hline
\end{tabular}}
\setlength{\tabcolsep}{1.4pt}
% \vspace{-.5cm}
\end{table}

%% file: tables/two_tables_enhance.tex
\begin{table}[t]
\setlength{\tabcolsep}{2pt}
% \scriptsize
% \footnotesize
\begin{center}
\caption{\textbf{SOTA results:} Quantitative evaluation with state-of-the-art methods on the deblurring dataset.}
\label{tab:results_deblurring}
\begin{tabular}{ll}
\hline
Method       & PSNR  \\ \hline
CNN-Baseline~\cite{hradivs2015convolutional} & 19.36 \\
Pix2Pix-HD~\cite{wang2018high}   & 19.89 \\
DE-GAN~\cite{souibgui2020gan}       & 20.37 \\
DocEnTr~\cite{souibgui2022docentr}      & 21.28 \\
\textbf{Ours}         & \textbf{23.58} \\ \hline
\end{tabular}
\end{center}
% \vspace{-0.5cm}
\end{table}

% \begin{SCtable}
% \setlength{\tabcolsep}{2pt}
% \centering
% % \scriptsize
% % \begin{minipage}[c]{0.33\textwidth}
% \caption{\textbf{Ablations of the degradations as pre-training objectives.} Results in document image binarization on DIBCO 2018 obtained by each pretext task in terms of PSNR.}
% \label{tab:ablation_enhance_objectives}
% % \end{minipage}\hfill
% % \begin{minipage}[c]{0.67\textwidth}
% \begin{tabular}{llll}
% \hline
% $\mathcal{L}_{mask}$ & $\mathcal{L}_{blur}$ & $\mathcal{L}_{noise}$ & PSNR  \\ \hline
% \xmark   & \xmark   & \xmark    & 18.75 \\
% \cmark   & \xmark   & \xmark   & 19.65 \\
% \xmark   & \cmark   & \xmark   & 18.98\\
% \xmark   & \xmark   & \cmark   & 19.82 \\
% \xmark   & \cmark  & \cmark   & 19.34 \\
% \cmark  & \xmark   & \cmark   & 19.45 \\
% \cmark  & \cmark  & \cmark   & \textbf{19.95} \\ \hline
% \end{tabular}
% \setlength{\tabcolsep}{1.4pt}
% % \end{minipage}
% \vspace{-0.5cm}
% \end{SCtable}

%% file: tables/dibco_table.tex
\begin{table*}[h]
\caption{\textbf{SOTA results.} Comparison of the proposed Text-DIAE compared to previous state-of-the-art approaches on the different DIBCO and H-DIBCO Benchmarks}
\label{tab:dibco_sota}
\resizebox{\textwidth}{!}{%
\begin{tabular}{l|cccccccccccccccc}
\hline
\multirow{3}{*}{Method} & \multicolumn{16}{c}{DIBCO Benchmarks} \\ \cline{2-17} 
 & \multicolumn{4}{c|}{2011} & \multicolumn{4}{c|}{2012} & \multicolumn{4}{c|}{2017} & \multicolumn{4}{c}{2018} \\
 & PSNR$\uparrow$ & FM$\uparrow$ & F$_{ps}$$\uparrow$ & \multicolumn{1}{c|}{DRD$\downarrow$} & PSNR$\uparrow$ & FM$\uparrow$ & F$_{ps}$$\uparrow$ & \multicolumn{1}{c|}{DRD$\downarrow$} & PSNR$\uparrow$ & FM$\uparrow$ & F$_{ps}$$\uparrow$ & \multicolumn{1}{c|}{DRD$\downarrow$} & PSNR$\uparrow$ & FM$\uparrow$ & F$_{ps}$$\uparrow$ & DRD$\downarrow$ \\ \hline
\cite{sauvola2000adaptive} & 15.60 & 82.10 & - & \multicolumn{1}{c|}{8.50} & 16.71 & 82.89 & 87.95 & \multicolumn{1}{c|}{6.59} & 14.25 & 77.11 & 84.1 & \multicolumn{1}{c|}{8.85} & 13.78 & 67.81 & 74.08 & 17.69 \\
\cite{kang2021complex} & 19.90 & \textbf{95.50} & - & \multicolumn{1}{c|}{1.80} & 21.37 & 95.16 & 96.44 & \multicolumn{1}{c|}{1.13} & 15.85 & 91.57 & 93.55 & \multicolumn{1}{c|}{2.92} & 19.39 & 89.71 & 91.62 & \textbf{2.51} \\
\cite{zhao2019document} & 20.30 & 93.80 & - & \multicolumn{1}{c|}{1.80} & 21.91 & 94.96 & 96.15 & \multicolumn{1}{c|}{1.55} & 17.83 & 90.73 & 92.58 & \multicolumn{1}{c|}{3.58} & 18.37 & 87.73 & 90.60 & 4.58 \\
\cite{souibgui2022docentr} & 20.81 & 94.37 & 96.15 & \multicolumn{1}{c|}{1.63} & 22.29 & 95.31 & 96.29 & \multicolumn{1}{c|}{1.60} & 19.11 & 92.53 & 95.15 & \multicolumn{1}{c|}{2.37} & 19.46 & 90.59 & 93.97 & 3.35 \\
\textbf{Ours} & \textbf{21.29} & 95.01 & \textbf{96.86} & \multicolumn{1}{c|}{\textbf{1.48}} & \textbf{23.66} & \textbf{96.52} & \textbf{97.04} & \multicolumn{1}{c|}{\textbf{1.10}} & \textbf{19.64} & \textbf{93.84} & \textbf{95.71} & \multicolumn{1}{c|}{\textbf{1.93}} & \textbf{19.95} & \textbf{91.32} & \textbf{94.44} & 3.21 \\ \hline
\end{tabular}

}
\end{table*}

%% file: tables/ablation_enhance_objectives.tex
\begin{SCtable}
\centering
% \scriptsize
% \begin{minipage}[c]{0.33\textwidth}
\caption{\textbf{Ablations of the degradations as pre-training objectives.} Results in document image binarization on DIBCO 2018 obtained by each pretext task in terms of PSNR.}
\label{tab:ablation_enhance_objectives}
% \end{minipage}\hfill
% \begin{minipage}[c]{0.67\textwidth}
\begin{tabular}{llll}
\hline
$\mathcal{L}_{mask}$ & $\mathcal{L}_{blur}$ & $\mathcal{L}_{noise}$ & PSNR  \\ \hline
\xmark   & \xmark   & \xmark    & 18.75 \\
\cmark   & \xmark   & \xmark   & 19.65 \\
\xmark   & \cmark   & \xmark   & 18.98\\
\xmark   & \xmark   & \cmark   & 19.82 \\
\xmark   & \cmark  & \cmark   & 19.34 \\
\cmark  & \xmark   & \cmark   & 19.45 \\
\cmark  & \cmark  & \cmark   & \textbf{19.95} \\ \hline
\vspace{-.4cm}
\end{tabular}
% \setlength{\tabcolsep}{1.4pt}
% \end{minipage}
\end{SCtable}

%% file: tex/conclusion.tex
% In this work, we have presented a Text-Degradation Invariant Auto-Encoder (Text-DIAE) framework designed for visual representation learning. 
This work demonstrates the capability of learning richer representations through pretext degradation tasks. Self-supervised learning can immensely boost the performance of text recognition and document image enhancement without any requirement of labeled data. 
Notably, we show that Text-DIAE does not share the limitations of contrastive or sequential approaches and is more effective at learning rich representations while seeing \textit{significantly} less data points. Extensive experimentation during fine-tuning demonstrate that Text-DIAE surpasses previous supervised and self-supervised state-of-the-art in handwritten text recognition and document image enhancement, while outperforming previous self-supervised approaches in scene-text recognition. 
We hypothesize that Text-DIAE performs complex variable reconstructions during pre-training, which helps to learn meaningful visual concepts from the latent representation space. 
We also provide the community the following insights to work on : 1) Designing new pretext tasks that are similar to downstream tasks.
2) The effect/trade-off of combination of various pretext tasks on the downstream tasks.
% 3) The search for an architectural pipeline that can solve multiple tasks.  
3) A need for a holistic approach to combine all the tasks into a single model.

%% file: appendix.tex
\appendix

\noindent
\section{Overview}
The main intuition behind the Text-DIAE framework has been the degradation invariant features learnt during pre-training without any manually annotated samples, which is fine-tuned for key downstream application tasks: Text-Recognition and Document Image Enhancement. In this supplementary material, we discuss and highlight some essential insights of Text-DIAE under the two sections~\ref{s:recognition} and ~\ref{s:enhancement}. 
% The supplementary material is organized as follows:
% ....
% \section{Implementation Details}

\section{Text Recognition}\label{s:recognition}
\subsection{Transformations}
We begin showing the augmentations employed to the input images used at the pre-training stage. We carefully select transformations suited for text in order to not disrupt the sequential information of characters found at the word level. 

\begin{figure}[h]
\centering
\includegraphics[width=\linewidth]{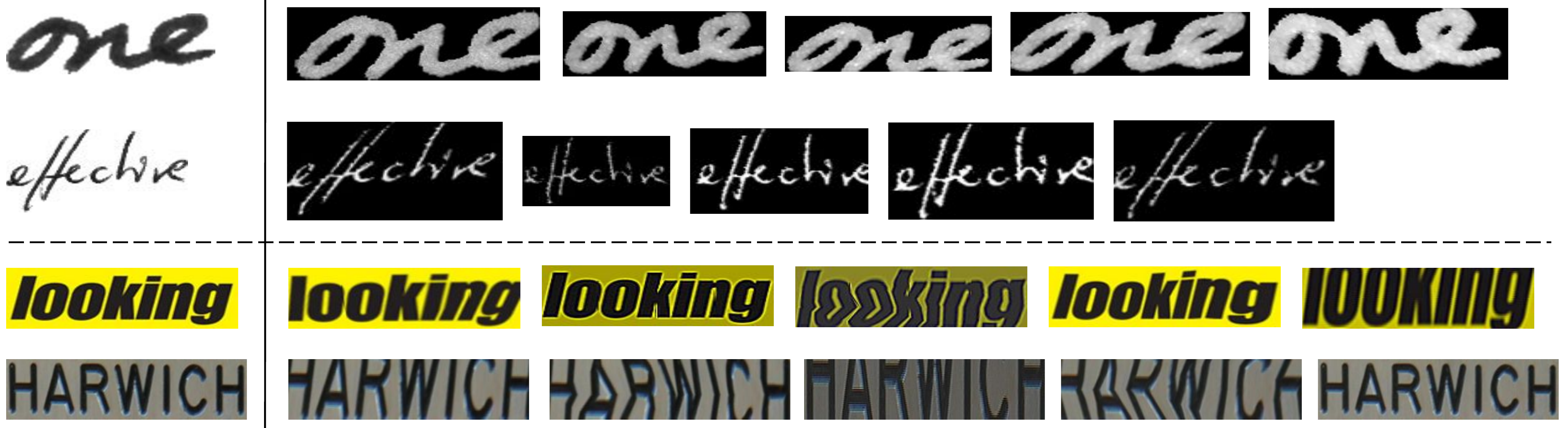}
\caption{\textbf{Transformations during training.} Image augmentations are carefully selected in order to not disrupt the sequentially nature of characters contained in a word.}
\label{fig:trans_da}
\end{figure}

Results of transformations employed can be seen in Figure~\ref{fig:trans_da}. The transformations used are: addition of Gaussian noise, shear, minor rotation, scaling and cropping upon some detected points given by a mask in order to not alter the characters in a significant manner.
Code regarding the data augmentation pipeline employed will be made public at \url{UponAcceptance}.

\setlength{\tabcolsep}{2pt}
\begin{table*}[t]
% \footnotesize
\begin{center}

\caption{\textbf{Implementation Details.} Implementations details of Handwritten and Scene Text Recognition. The acronyms STR and HTR stands for Scene Text Recognition and Handwritten Text Recognition respectively.} 
\label{tab:model_configs}
\resizebox{\textwidth}{!}{
\begin{tabular}{llll}
\hline\noalign{\smallskip}
ConFigure & Pre-Training & Fine tuning & Scratch \\
\noalign{\smallskip}
\hline
\noalign{\smallskip}
optimizer & AdamW & Adam & Adam \\
learning rate & 1.5 $e^{-4}$ & 1 $e^{-4}$ & 1.5 $e^{-5}$ \\
weight decay & 0.05 & 0.05 & 0.05 \\
optimizer momentum & $\beta_1$, $\beta_2$=0.9, 0.95 & $\beta_1$, $\beta_2$=0.9, 0.95 & $\beta_1$, $\beta_2$=0.9, 0.95 \\
batch size & 64 (HTR) / 192 (STR) & 64 (HTR) / 256 STR & 64 (HTR) / 256 STR \\
learning rate schedule & cosine decay & cosine decay & cosine decay \\
warmup epochs & 3 & 3 & 10 \\
training epochs & 100 (HTR) / 2 (STR) & 600 (HTR) / 10 (STR) & 600 (HTR) / 10 (STR) \\
\hline
\end{tabular}}
\end{center}
\vspace{-0.5cm}
\end{table*}
\setlength{\tabcolsep}{1.4pt}

\begin{figure*}[h]
\centering
\includegraphics[width=\linewidth]{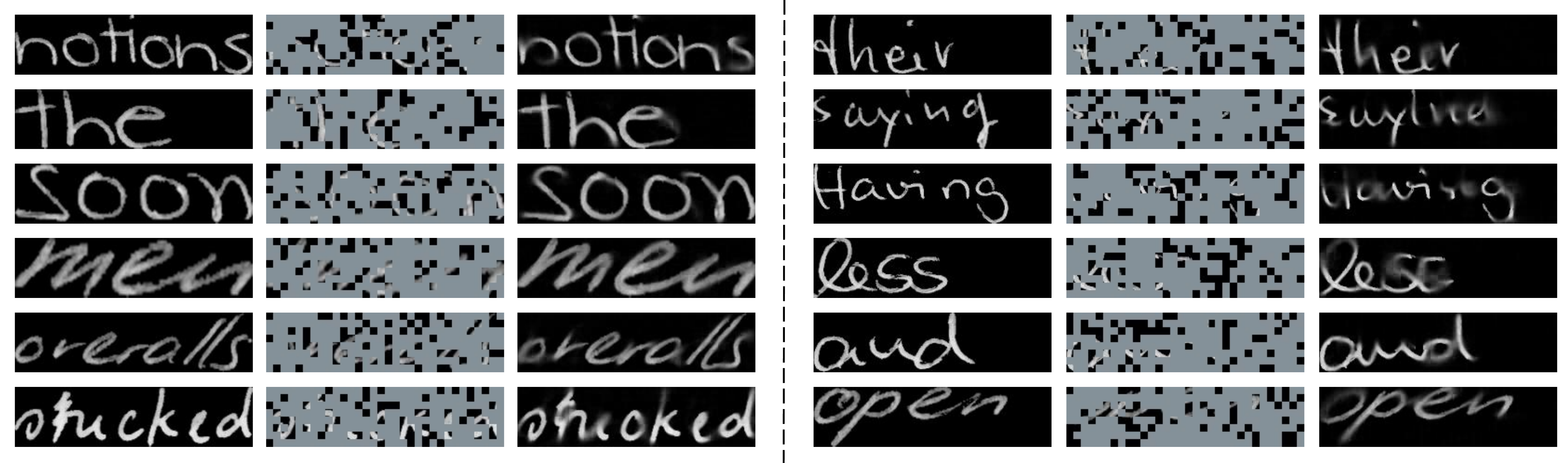}
\caption{\textbf{Reconstruction of Handwritten Text.} Qualitative samples of reconstructed input images. The model may have seen the word, but it has not seen the images at training. The reconstruction is performed after employing masking as pretext task during pre-training.}
\label{fig:htr_masking}
\end{figure*}

\subsection{Datasets}
In this work, we make use of the following public datasets to train models in handwriting and scene text recognition.
\begin{itemize}
    \item \textbf{IAM}~\cite{marti2002iam}: It is a handwritten English text dataset. It contains 657 different writers, and it is partitioned into writer independent training, validation and test. Addtionally, this dataset is comprised of 74,805 fully segmented words.
    \item \textbf{CVL}~\cite{kleber2013cvl}: Dataset of handwritten text in English language. It was written 
    310 different writers, partitioned into writer independent training and test sets. In total, 27 writers wrote 7 texts and the remaining 283 writers wrote 5 texts.
    \item \textbf{IIIT5K-words} (IIIT5K)~\cite{mishra2012scene}: This is a dataset made of 2000 training and 3000 testing cropped scene text images from the Internet. Often employed in recognition and scene text retrieval tasks.
    \item \textbf{ICDAR-2013} (IC13)~\cite{karatzas2013icdar}: It is a dataset that contains 848 training and 1015 testing cropped scene text in focused environments, where the text was relevant. Often employed in text detection and recognition.
    \item \textbf{Modified-SynthText}: This dataset is a modified version of~\cite{gupta2016synthetic}. It was employed by~\cite{gomez2018single} to train a scene-text retrieval model. It contains 4M cropped scene text images which were generated synthetically. This dataset is smaller than the original one and it avoids random flipping of text images and rotated text that can break the sequential order of characters. 
    % In this work, we use this dataset to pre-train the scene text encoder.

\end{itemize}
The pre-training datasets are domain-dependant, in handwritten text we used IAM and CVL. In scene text we used the Modified-SynthText as well as  the training sets from ICDAR13~\cite{karatzas2013icdar} and IIIT5K~\cite{mishra2012scene}.
% What to include?
\subsection{Evaluation metrics}

\noindent
\textbf{Character Error Rate (CER).} CER is used to compare the similarity of two strings at character level. Given two stings $s_1$ and $s_2$, where $s_2$ is the reference (ground truth):
\begin{equation}
    CER(s_1,s_2) = \frac{S+D+I}{N}
\end{equation}
where $S$ is the number of substitutions, $D$ is the number of deletions and $I$ is the number of insertions that are applied in $s_1$ to obtain $s_1 = s_2$. Noting that,  $N$ is the number of characters in $s_2$.

\noindent
\textbf{Edit Distance at 1 (ED1).} It represents the accuracy of a prediction up to a edit distance of 1. This implies that a prediction is correct compared to the ground truth if only \textit{one} substitution, deletion or insertion is required for both strings to match.

\subsection{Implementation details}

In order to provide the reader with reproducibility of the presented work, the implementation details of the models showcased for Handwritten and Scene Text Recognition are shown and summarized in Table~\ref{tab:model_configs}. 

During pre-training we deploy an encoder with 6 layers and 8 attention heads to encode the input, with a dimension of 768. We used this same number of layers and attention heads for the decoder, with a dimension of 512 in text recognition and 768 for document enhancement.  
At masking, each input image (with size $64\times256\times3$ for text recognition and $256\times256\times3$ for document enhancement) is divided into a set of patches with size $8\times8\times3$. Similarly as~\cite{he2021masked}, we employ random masking of 75\% of the patches. To add blur, we add average blur with random kernel sizes between 1 and 15. In order to add background noise, we add weighted contrasting backgrounds from different text documents.  The pre-training was done for 2 epochs in scene text and 100 epochs for the handwritten domain.

In the fine tuning stage, we use the same encoder (pre-trained) with a different decoder of 6 layers, 8 attention heads and dimension of 768.

\subsection{Qualitative Insights}

\begin{figure*}[h]
\centering
\includegraphics[width=\linewidth]{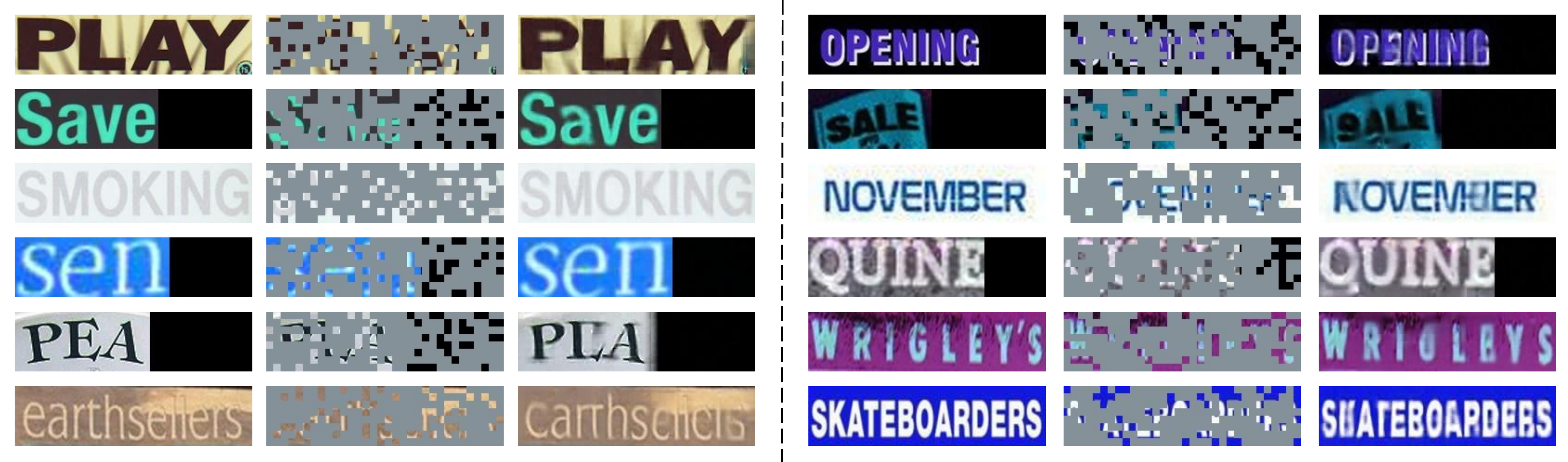}
\caption{\textbf{Reconstruction of Scene Text.} Qualitative samples of reconstructed input images. The model may have seen the word, but it has not seen the images at training. The reconstruction is performed after employing masking as pretext task during pre-training.}
\label{fig:str_masking}
\end{figure*}

\begin{figure*}[h]
\centering
\includegraphics[ width=\linewidth]{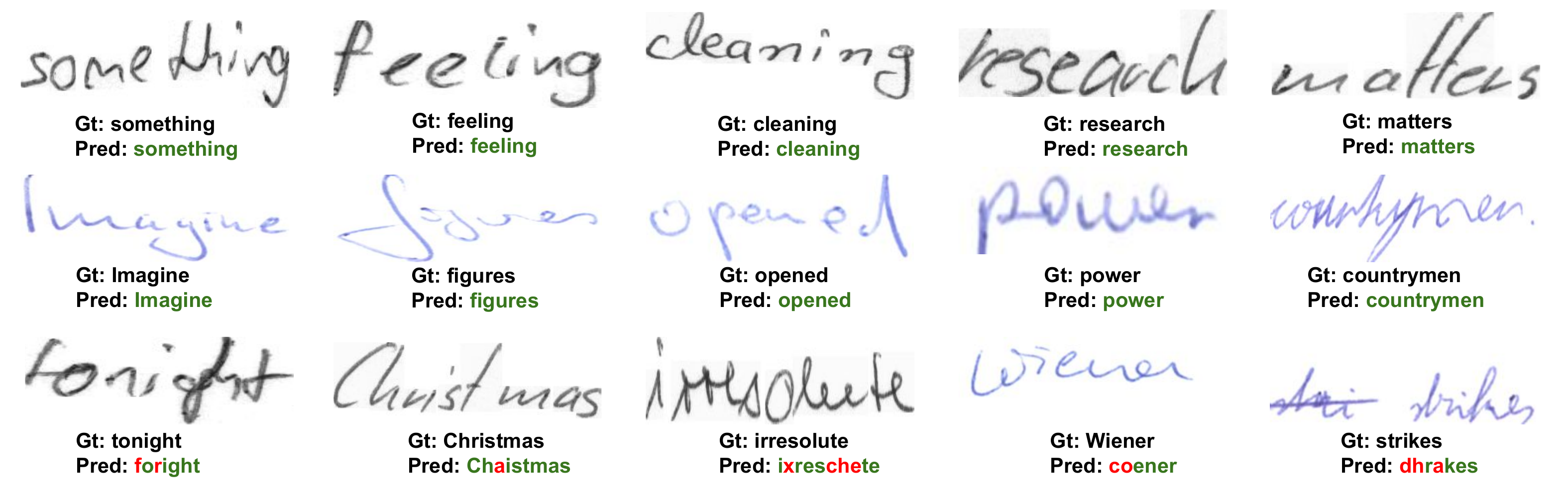}
\caption{\textbf{Handwritten Text Recognition.} Qualitative samples of attaching a decoder and fine-tuning an encoder to perform recognition of IAM (black font) and CVL (blue font). Correctly predicted samples are displayed in green, while mistakes are shown in red. Failure cases are depicted on the last row.}
\label{fig:text_rec_qualitative}
\end{figure*}

In Figure~\ref{fig:htr_masking} we show qualitative samples of the original, masked and reconstructed images at pre-training. The images correspond to the validation set and the model have not seen them. In Figure~\ref{fig:htr_masking}, we focus on the pre-text task (masking - 75\%) that yields the biggest improvement on the downstream task. We note that at pre-training the model learns a distribution of characters (English) that helps to solve the masking task. It is worth noting the capability of keeping similar font types and applying a padding in short words. In harder cases, such as in the words "stucked", "saying" and "having" the model still reconstructs blurry or wrong characters. However, we have to note the complexity of this task, which is very demanding even for humans.

Similarly, we show qualitative samples of reconstructing images at pre-training by masking an input scene-text cropped word in Figure~\ref{fig:str_masking}. The model has not seen this images before. The model learns to properly reconstruct input images, while properly keeping the background color and specific font style. Some failure cases are shown while reconstructing "pea" and "earthsellers".

We show in Figure~\ref{fig:text_rec_qualitative} the recognized text in the test sets of IAM and CVL datasets. Despite the difficulty of some samples, the model achieves state-of-the-art in IAM. Failure cases depict problematic handwritten styles to be recognized, e.g. "tonight" and "Christmas". Some characters are mistakenly predicted, thus causing errors in the next sequence of characters, such in "irresolute" and "wiener". Some errors come from difficult cropped samples such as "strikes", in which the sample starts with a strikethrough word.

\section{Document Image Enhancement}\label{s:enhancement}

\subsection{Datasets}

% \noindent
\begin{itemize}
    \item \textbf{DIBCO and H-DIBCO datasets}: These datasets were introduced in the Document Image Binarization challenges that took place since 2009. In this work we have mainly chosen two DIBCO(2011~\cite{Pratikakis2011icdar} and 2017~\cite{pratikakis2017icdar2017}) and two H-DIBCO(2012~\cite{pratikakis2012icfhr} and 2018~\cite{Pratikakis2018icfhr}) benchmarks for final validation on the document image binarization task. All the image samples are mainly historical documents that has undergone several distortions like bleed, show-through, smears, fading and so on. The DIBCO 2011 and 2017 contained 16 and 20 samples respectively. While the H-DIBCO 2012 and 2018 contained 14 and 10 instances respectively.
    
    \item \textbf{Blurry Document Images}: The training dataset contains 4000 training images and 932 images for validation of resolution 300x300 patches. This data was used in ~\cite{souibgui2020gan} and it is originally a subset from the dataset proposed in~\cite{hradivs2015convolutional}. Every instance is extracted from a different document image and each blur kernel used is exclusive.

\end{itemize}
 For pre-training we used 203,576 unlabeled document image samples. These samples were taken from the historical document DIBCO benchmarks (2009, 2010, 2013, 2014, 2015, 2016)~\cite{pratikakis2016icfhr2016}, Publaynet~\cite{zhong2019publaynet}, Palm-leaf~\cite{burie2016icfhr2016} and IAM Handwriting database.
\subsection{Evaluation metrics}

\noindent
\textbf{Peak signal-to-noise ratio (PSNR).} PSNR helps to establish a pixel-wise validation and measures the effectiveness of document image enhancement approaches in terms of visual quality. It computes the ratio between the maximum possible value of the signal(image) to the amount of noise (distortion) that affects the quality.  The higher the value, the more similar are the two images which in turn assures that the reconstructed image (binarized or deblurred) has a better quality. MAX is the maximum possible pixel value of the image (eg. MAX is 255 for pixels represented as 8 bits per sample) . Given two $M*N$ images, they can be fomulated as shown in eqn.~\ref{eq:mse_psnr}.

\begin{equation}
\begin{split}
    M S E=\frac{\sum_{M, N}\left[I_{1}(m, n)-I_{2}(m, n)\right]^{2}}{M * N}
    \\ P S N R=10 \log _{10}\left(\frac{MAX^{2}}{M S E}\right)
\end{split}
\label{eq:mse_psnr}
\end{equation}
\\

\noindent
\textbf{F-Measure(FM).} The F-measure gives the harmonic mean between the precision (P) and recall (R) scores. Precision determines the number of positive predictions, while recall measures the ability to find the positive predictions in a binary classifier. Here, in binarization problem, FM computes the accurate prediction of the white background and the black foreground (text) between a binarized sample and the ground-truth(GT) sample. FM is formulated as shown in eqn.~\ref{eq:fmeasure}.     

\begin{equation}
\begin{split}
   F M=\frac{2 * \text {Recall} * \text {Precision}}{\text {Recall}+\text {Precision}}
\end{split}
\label{eq:fmeasure}
\end{equation}
\\

\noindent
\textbf{Pseudo F-Measure ($F_{ps}$).} The Pseudo F-Measure was first introduced for document image binarization task in~\cite{ntirogiannis2012performance} as it proposed a normalization procedure on the weights of the GT image(both background and foreground) for a lower penalization. 
\\

\noindent
\textbf{Distance Reciprocal Distortion (DRD).} DRD was mainly used to measure visual distortion for all pixels in binary document images as proposed in~\cite{lu2004distance}. It is computed at first kth flipped pixel using a normalized weight matrix (5x5) $W_{norm}$ defined in the same paper~\cite{lu2004distance}. The formulation is done as shown in eqn.~\ref{eq:drd}. To calculate the overall DRD score between the binarized resultant image $B_{k}(x, y)$ and the GT image $G T_{k}(i, j)$  , it is basically to sum up the DRD for $N$ number of flipped pixels starting from $k=1$ and $NB_{nu}$ refers to the number of non-uniform blocks as formulated in eqn.~\ref{eq:drd_final}.

\begin{equation}
\begin{split}
   D R D_{k}=\sum_{i=-2}^{2} \sum_{j=-2}^{2}\left|G T_{k}(i, j)-B_{k}(x, y)\right| \times W_{norm}(i, j)
\end{split}
\label{eq:drd}
\end{equation}

\begin{equation}
    \begin{split}
        D R D=\frac{\sum_{k=1}^{N} D R D_{k}}{NB_{nu}}
    \end{split}
    \label{eq:drd_final}
\end{equation}

\subsection{Implementation details}

The implementation details of the models showcased for Document Image Enhancement tasks (binarization and deblurring) are shown and summarized in Table~\ref{tab:model_configs_enhancement}. The Xavier-Glorot uniform~\cite{glorot2010understanding} was used to initialize the transformer blocks, as implemented in the original ViT~\cite{dosovitskiy2020image}. The ViT-Base model variant was used as the main model baseline with the patch size $8$ and Random resized Crop was used as the default applied transformation for all the images during pre-training.     
The pre-training was done for 100 epochs.
\setlength{\tabcolsep}{2pt}
\begin{table}[h]
\centering
\caption{\textbf{Training settings:} Text-DIAE model configurations and hyperparameters adapted for Document Image Enhancement tasks.}
\label{tab:model_configs_enhancement}
\begin{tabular}{lcc}
\hline
ConFigure                & \multicolumn{1}{l}{Pre-Training} & \multicolumn{1}{l}{Fine tuning} \\ \hline
optimizer              & AdamW                            & AdamW                           \\
learning rate          & 1.5e-4                           & 1.5e-4                          \\
weight decay           & 0.05                             & 0.05                            \\
optimizer momentum     & (0.9,0.95)                       & (0.9,0.99)                      \\
batch size             & 64                               & 64                              \\
learning rate schedule & cosine                           & cosine                          \\
warmup epochs          & 3                                & 15                              \\
training epochs        & 50                               & 100                             \\ \hline
\end{tabular}
\end{table}
\setlength{\tabcolsep}{1.4pt}

\subsection{Qualitative Insights}
The visual appearance of a historical document has been negatively affected with time by different forms of degradation. Recovering the document with its foreground text and information is the key challenge in document image enhancement. In this section, we will discuss some qualitative highlights of our proposed Text-DIAE model applied to document image binarization. We shall also cover some analysis on the task of document deblurring mainly on camera-captured document images affected by motion, out-of-focus blurs etc. 

\begin{figure*}[h]
\centering
\setlength{\tabcolsep}{0.31em}
\begin{tabular}{c c c c}
     \toprule
     \scriptsize{Original Input} & \scriptsize{DocEnTr~\cite{souibgui2022docentr}} & \scriptsize{\textbf{Ours}} & \scriptsize{Ground Truth} \\
     \midrule
     \includegraphics[width=0.23\linewidth]{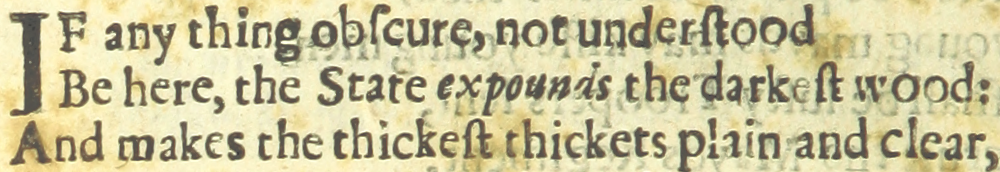} &  \includegraphics[width=0.23\linewidth]{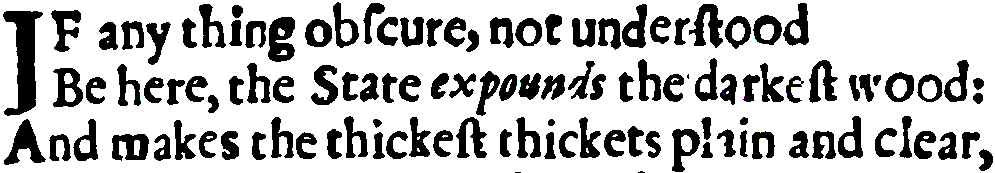} & \includegraphics[width=0.23\linewidth]{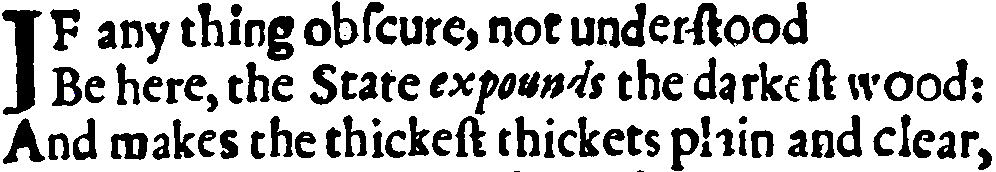} &
     \includegraphics[width=0.23\linewidth]{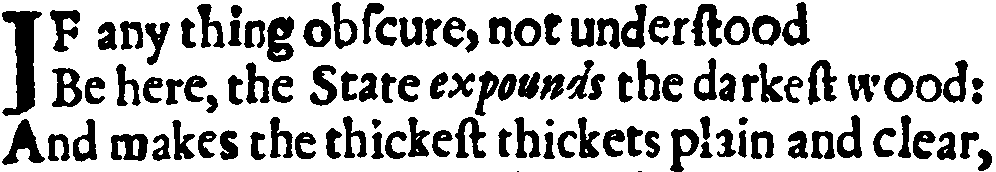} \\
     \midrule 
     \begin{tikzpicture}
  \node (ocr1) [text width=0.21\textwidth, font=\fontsize{7pt}{7pt}\selectfont,
  align=left] at (0,0) {
  \textbf{OCR output:} ``{\fontfamily{qcr}\selectfont 
    \color{Red}{J}\color{Green}{anythingob}\color{Red}{f}\color{Green}{cure,\\notunder}\color{Red}{f}\color{Green}{tood}\color{Red}{,,., ... ,}\color{Green}{Be here, the State expounds the darke}\color{Red}{f}\color{Green}{t wood: And makes the thicke}\color{red}{f}\color{Green}{t thickets p}\color{Red}{!}\color{Green}{ain}\color{Red}{-}\color{Green}{and clear,}}''
 };
\end{tikzpicture} &  \begin{tikzpicture}
  \node (ocr1) [text width=0.21\textwidth, font=\fontsize{7pt}{7pt}\selectfont,
  align=left] at (0,0) {%
  \textbf{OCR output:} ``{\fontfamily{qcr}\selectfont 
  \color{Red}{I}\color{Green}{any thing ob}\color{Red}{f}\color{Green}{cure,notunder}\\ \color{Red}{f}\color{Green}{tood Be here, the State expounds the darke}\color{Red}{f}\color{Green}{t wood: A}\color{Red}{o}\color{Green}{d makes}\color{Red}{c}\color{Green}{he thicke}\color{red}{f}\color{Green}{t thickets p}\color{Red}{!t}\color{Green}{in}\color{Green}{and clear,}}''
 };
 
\end{tikzpicture} & \begin{tikzpicture}
  \node (ocr1) [text width=0.21\textwidth, font=\fontsize{7pt}{7pt}\selectfont,
  align=left] at (0,0) {%
  \textbf{OCR output:} ``{\fontfamily{qcr}\selectfont 
  \color{Green}{I}\color{Red}{j}\color{Green}{any thing ob}\color{Red}{f}\color{Green}{cure,not under}\color{Red}{f}\color{Green}{tood Be here, the State expounds the darke}\color{Red}{f}\color{Green}{t wood: And makes}\color{Red}{c}\color{Green}{he thicke}\color{red}{f}\color{Green}{t thickets pl}\color{Red}{t}\color{Green}{in}\color{Green}{and clear,}}''
 };
\end{tikzpicture} &
     \begin{tikzpicture}
  \node (ocr1) [text width=0.21\textwidth, font=\fontsize{7pt}{7pt}\selectfont,
  align=left] at (0,0) {%
  \textbf{OCR output:} ``{\fontfamily{qcr}\selectfont 
  \color{Green}{F}\color{Green}{any thing ob}\color{Red}{f}\color{Green}{cure,not under}\color{Red}{f}\color{Green}{tood Be here, the State expounds the darke}\color{Red}{f}\color{Green}{t wood: And makes}\color{Red}{c}\color{Green}{he thicke}\color{red}{f}\color{Green}{t thickets p}\color{Red}{!}\color{Green}{ain}\color{Green}{and clear,}}''
 };
\end{tikzpicture} \\
     \midrule
     \scriptsize{CER: 17.97} &  \scriptsize{CER: 13.28} & \scriptsize{CER: \textbf{11.50}} & \scriptsize{CER: 10.94} \\
     \bottomrule
     \end{tabular}
\caption{\textbf{Qualitative analysis of binarized samples for OCR.} The document image on the left refers to the originally captured image, followed by the DocEnTr~\cite{souibgui2022docentr} binarized result, and the Text-DIAE binarized result, and the ground-truth image in the last column. The correctly predicted OCR output is shown in ”Green” font while the inaccurate ones are depicted in ”Red” and recognition performance in terms of CER.} 
\label{fig:ocr_dibco}
\end{figure*}

\begin{figure*}[!t]
% \begin{center}
\centering
  
 \begin{tabular}{ccc}
 
    \includegraphics[width=0.30 \linewidth, height=20mm]{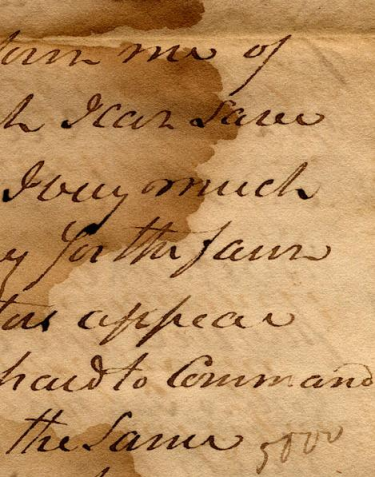} &
    \includegraphics[width=0.30 \linewidth, height=20mm]{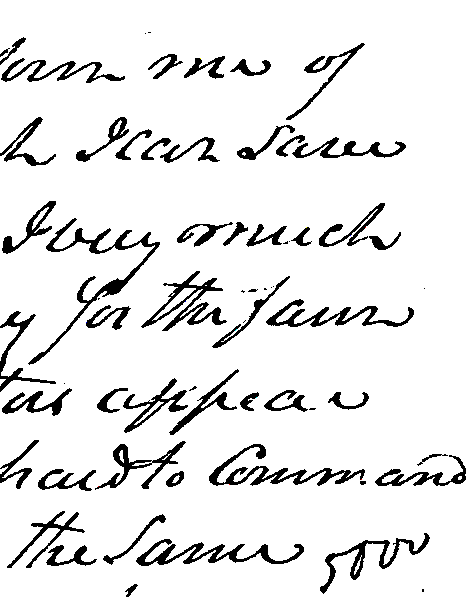}&
    \includegraphics[width=0.30 \linewidth, height=20mm]{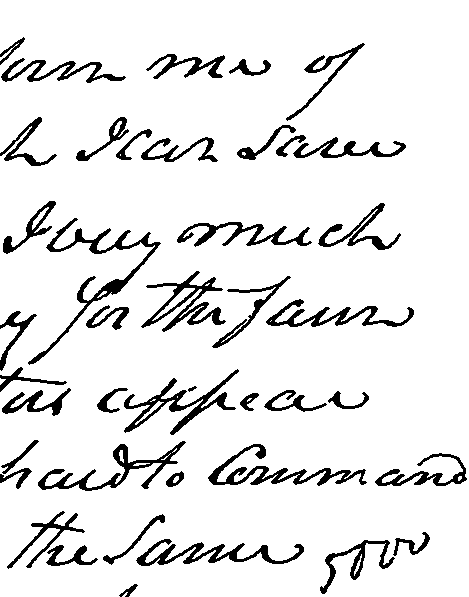}
    \\ \noalign{\smallskip} 
    
    \includegraphics[width=0.30 \linewidth, height=20mm]{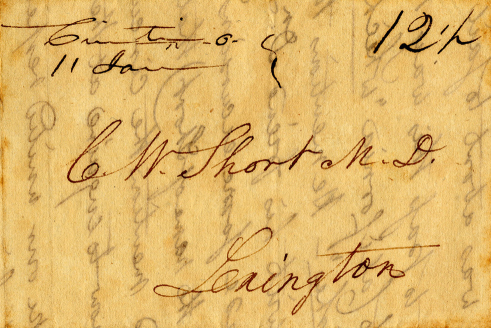} &
    \includegraphics[width=0.30 \linewidth, height=20mm]{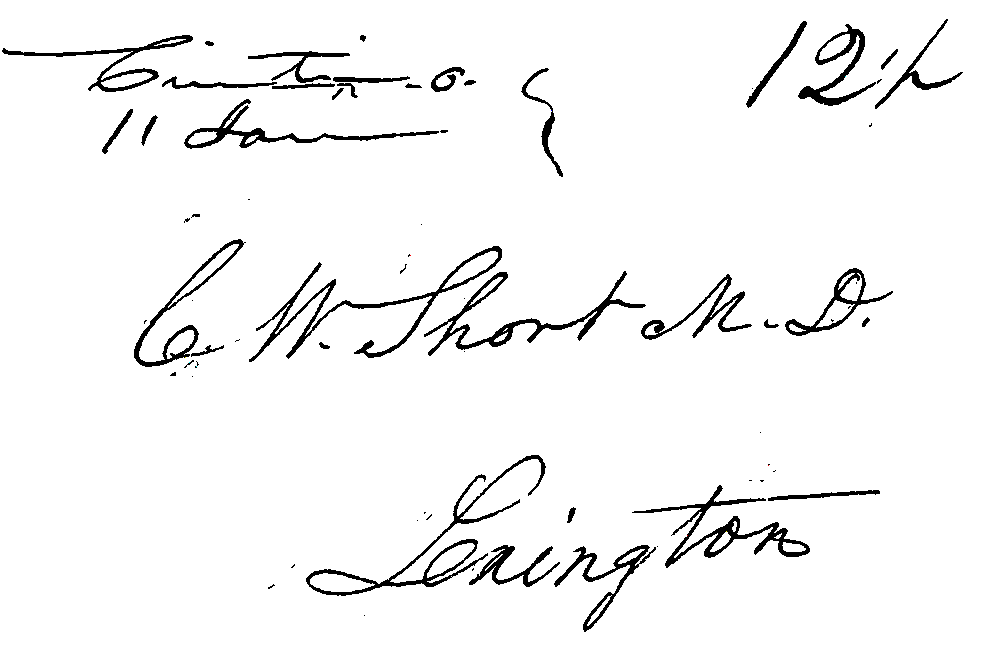}&
    \includegraphics[width=0.30 \linewidth, height=20mm]{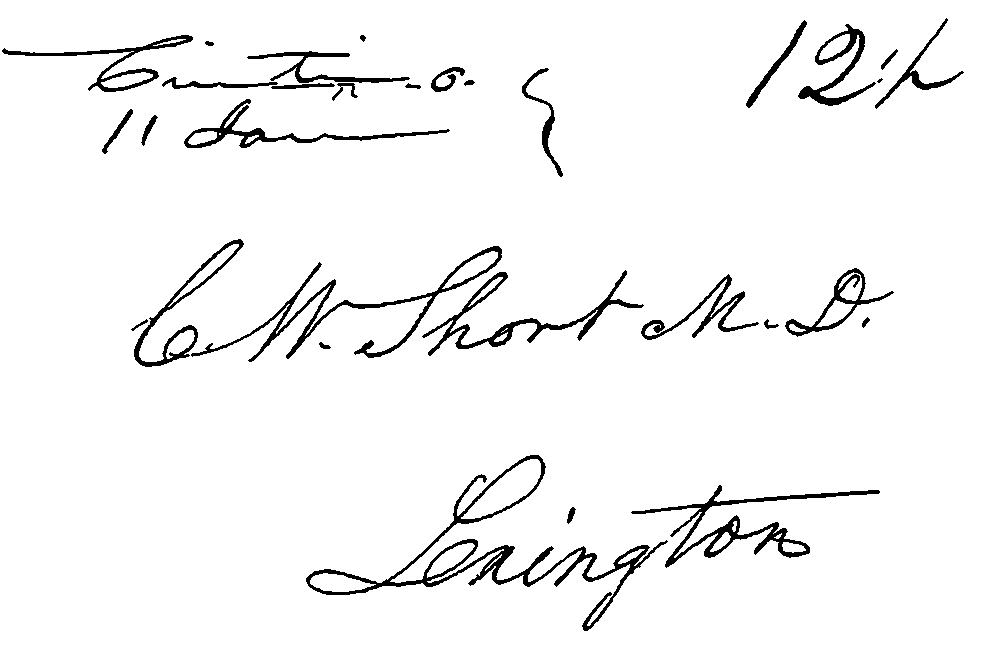}
    \\ \noalign{\smallskip}    
    
    \includegraphics[width=0.30 \linewidth, height=20mm]{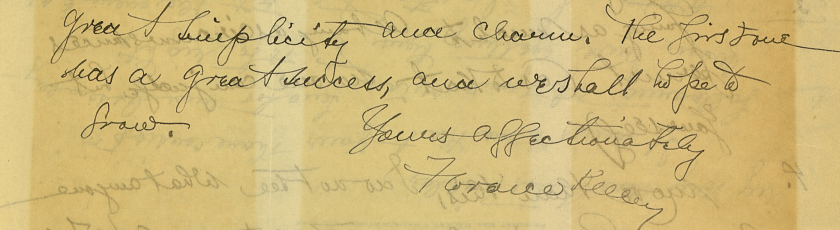} &
    \includegraphics[width=0.30 \linewidth, height=20mm]{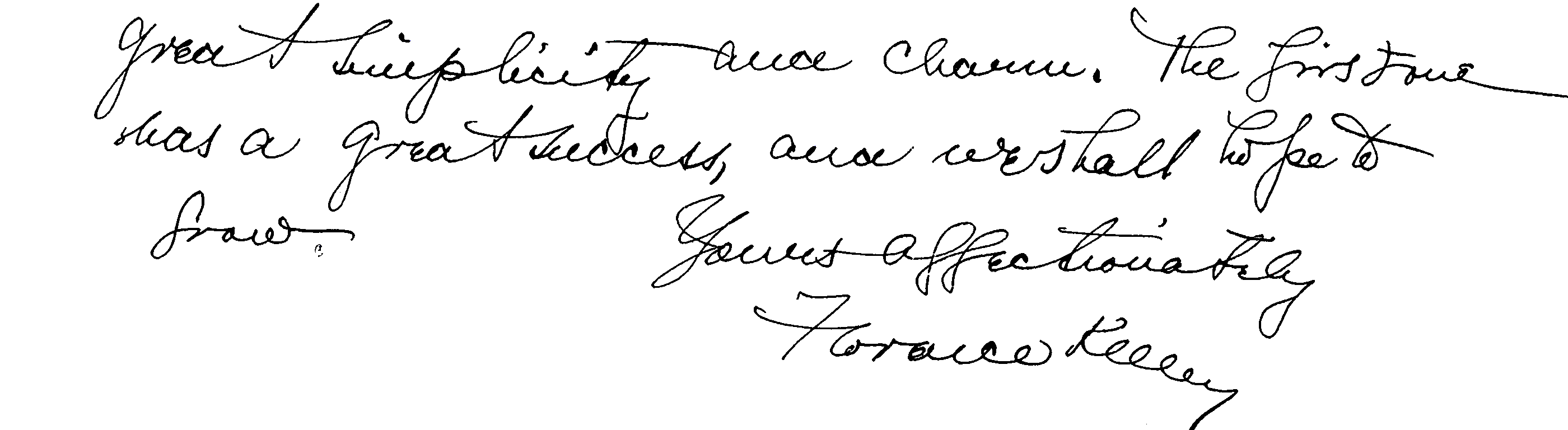}&
    \includegraphics[width=0.30 \linewidth, height=20mm]{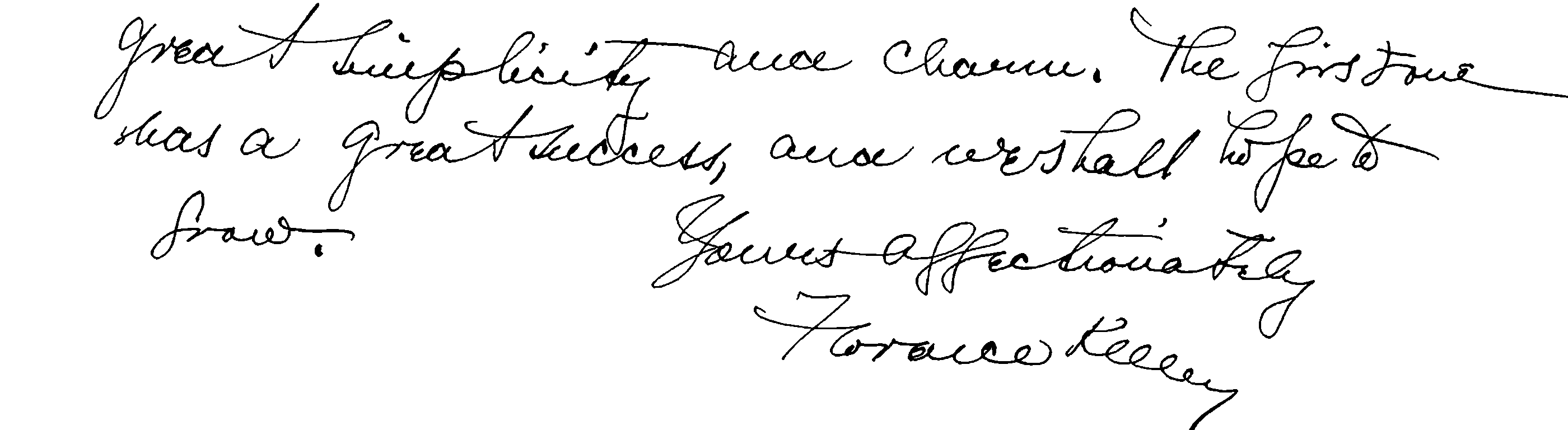}
    \\ \noalign{\smallskip}  
    
    \includegraphics[width=0.30 \linewidth, height=20mm]{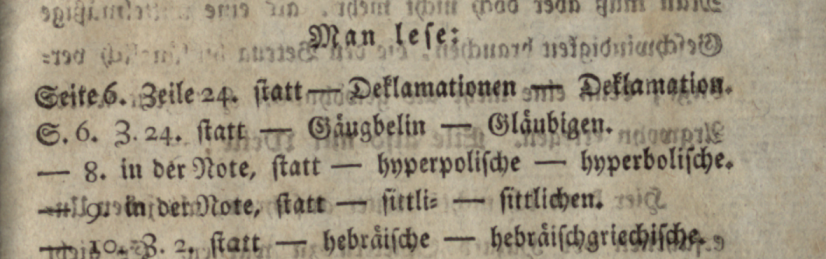} &
    \includegraphics[width=0.30 \linewidth, height=20mm]{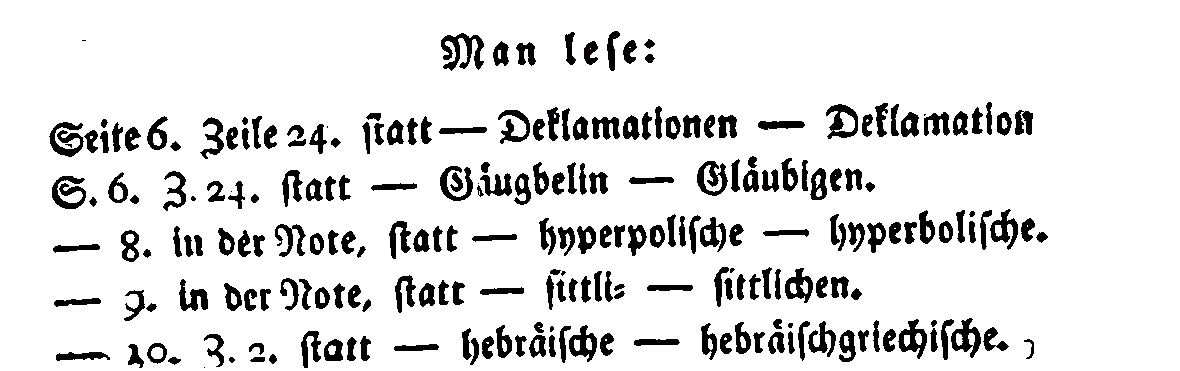}&
    \includegraphics[width=0.30 \linewidth, height=20mm]{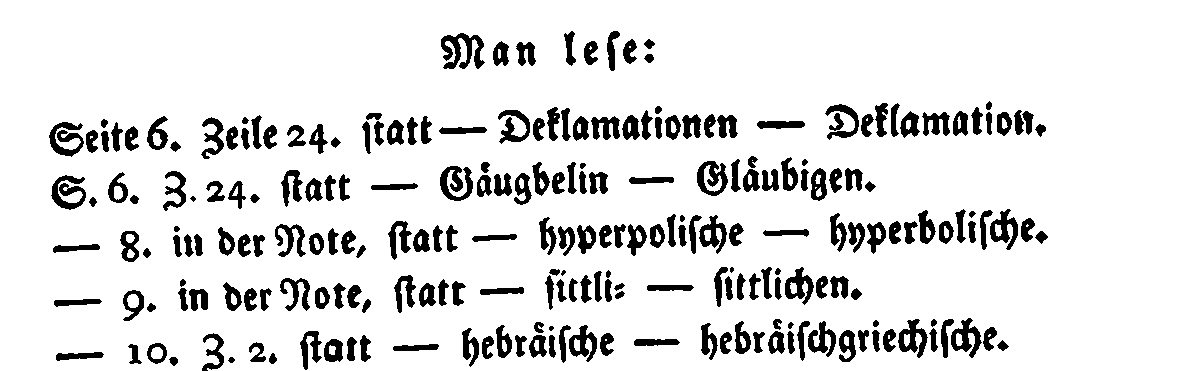}
    \\ \noalign{\smallskip}  
    
    \includegraphics[width=0.30 \linewidth, height=20mm]{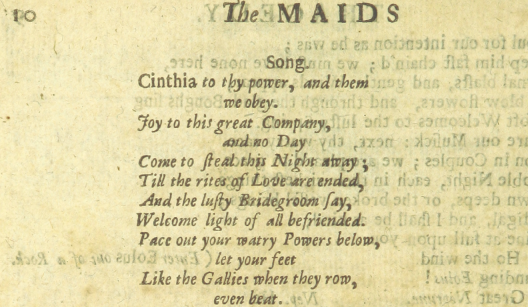} &
    \includegraphics[width=0.30 \linewidth, height=20mm]{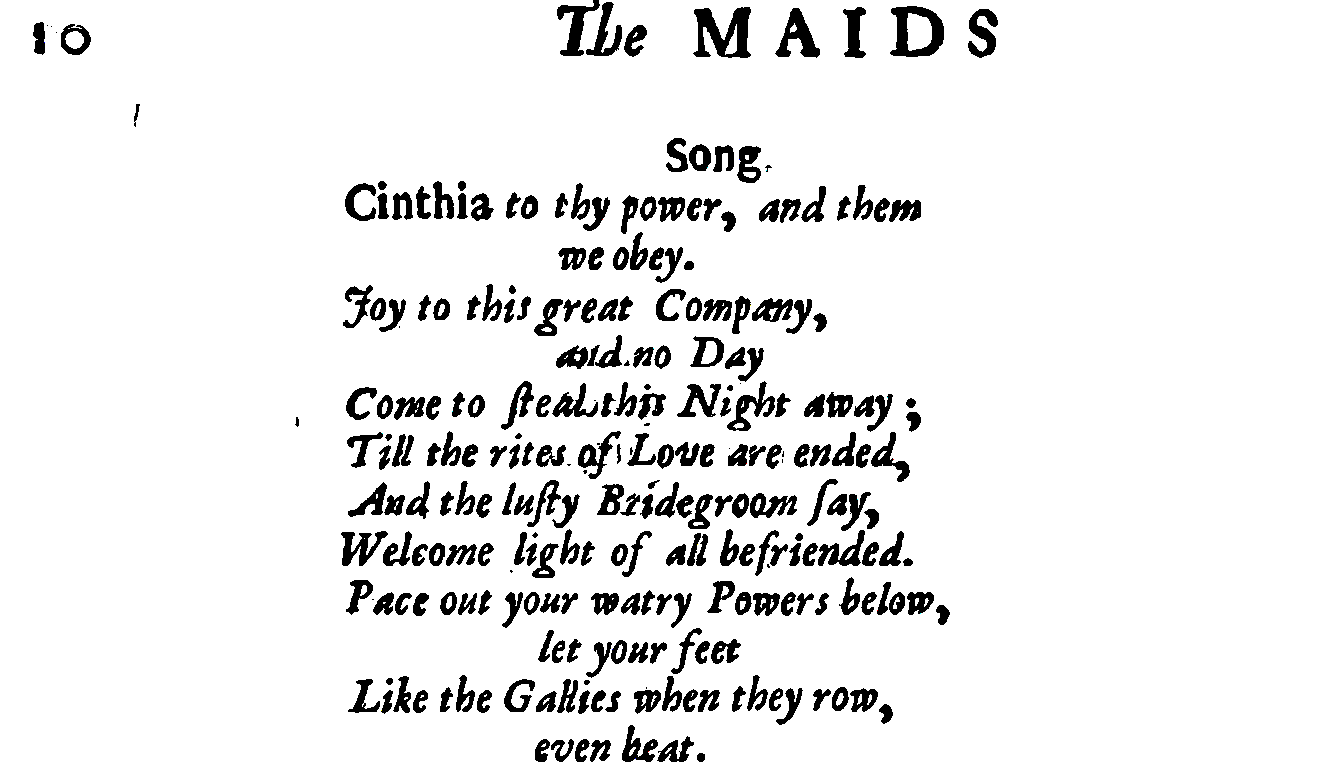}&
    \includegraphics[width=0.30 \linewidth, height=20mm]{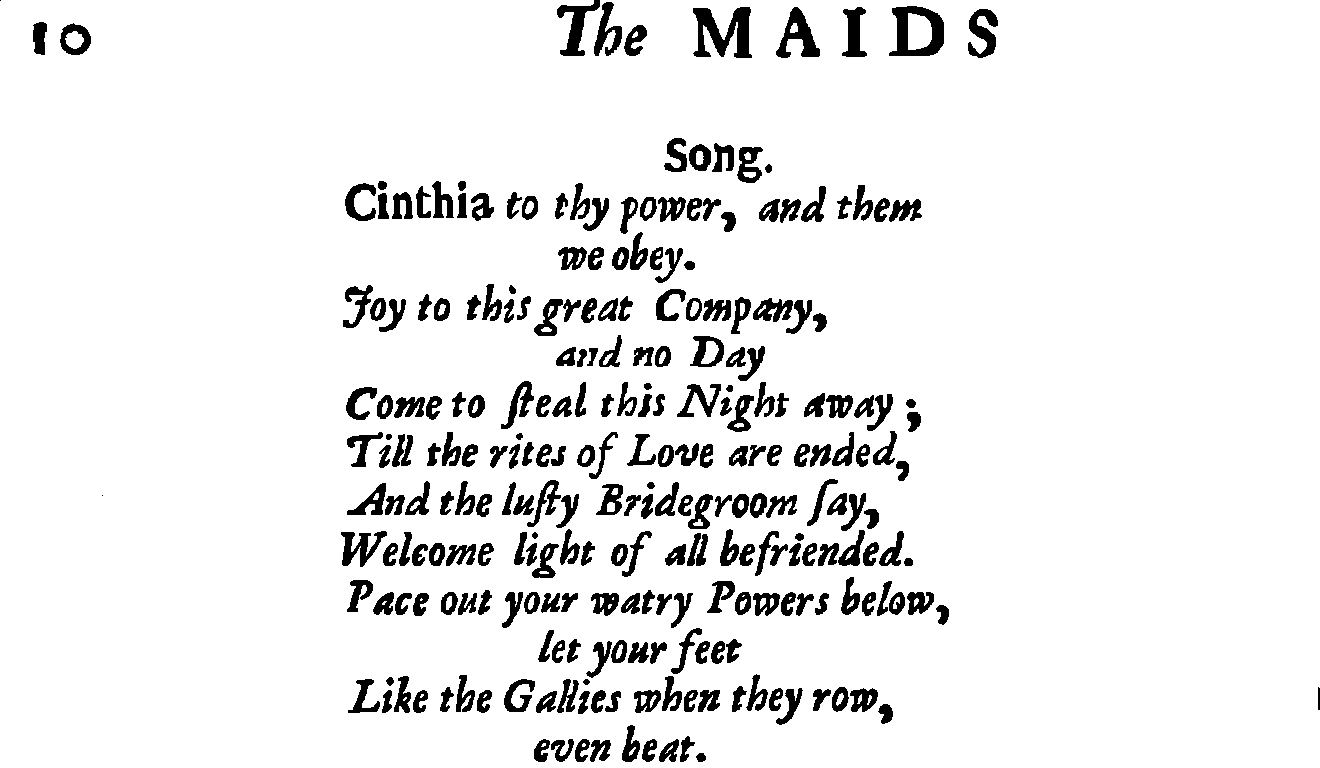}
    \\ \noalign{\smallskip}  
    
    \includegraphics[width=0.30 \linewidth, height=20mm]{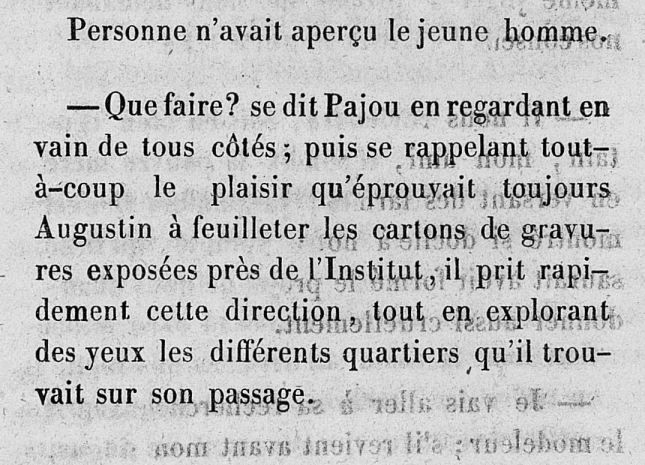} &
    \includegraphics[width=0.30 \linewidth, height=20mm]{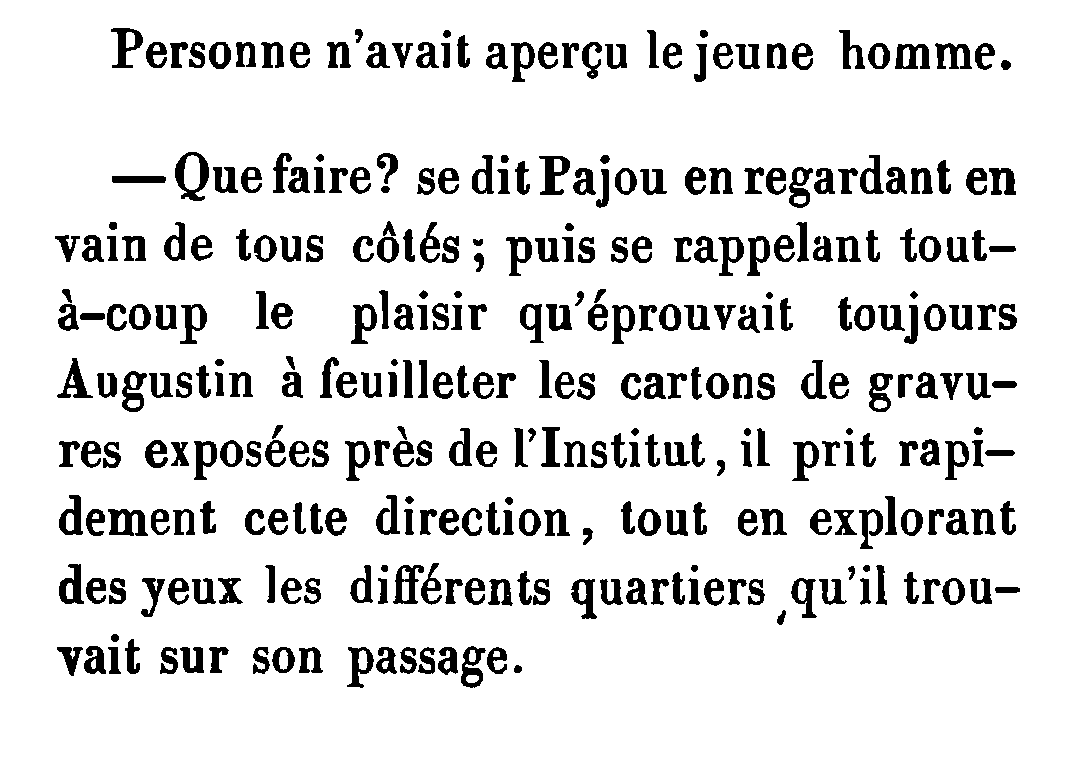}&
    \includegraphics[width=0.30 \linewidth, height=20mm]{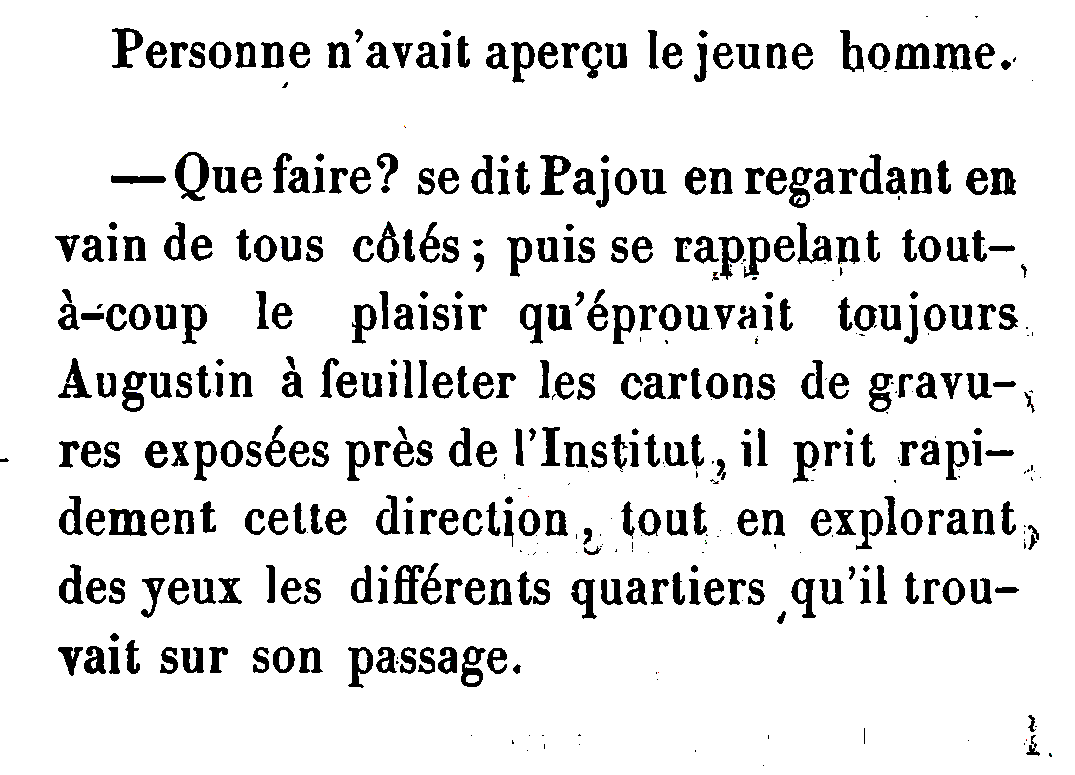}
    \\ \noalign{\smallskip}

    \includegraphics[width=0.30 \linewidth, height=20mm]{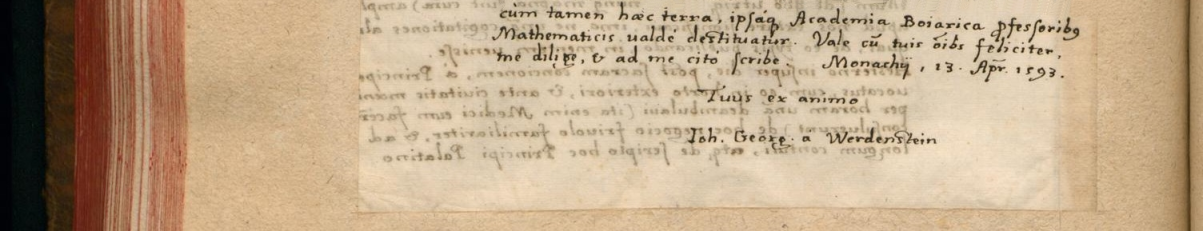} &
    \includegraphics[width=0.30 \linewidth, height=20mm]{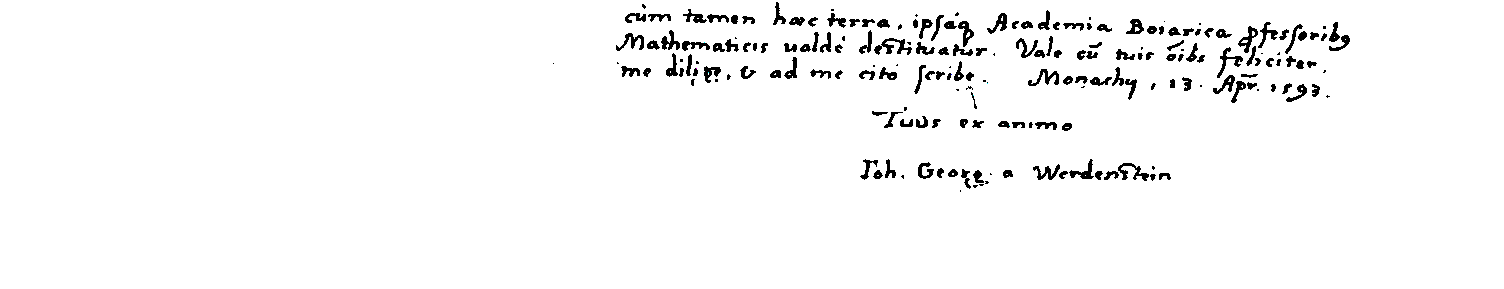}&
    \includegraphics[width=0.30 \linewidth, height=20mm]{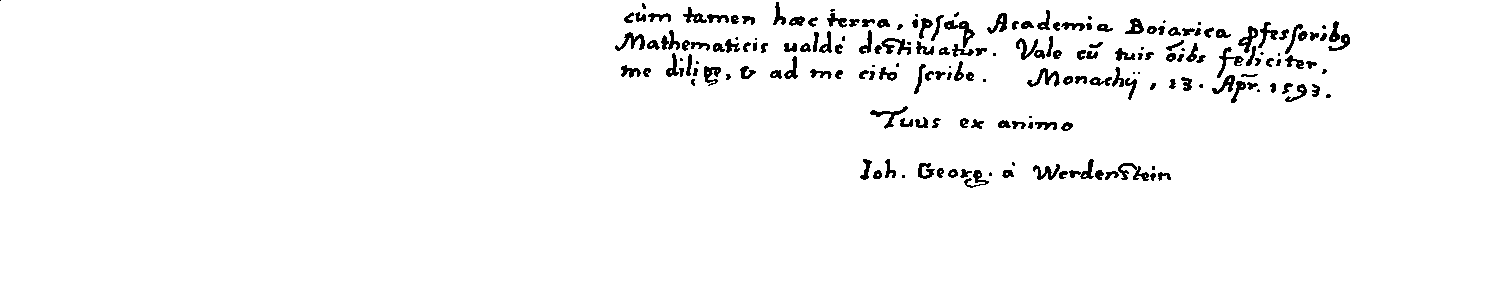}
    \\ \noalign{\smallskip}     
    
    \includegraphics[width=0.30 \linewidth, height=20mm]{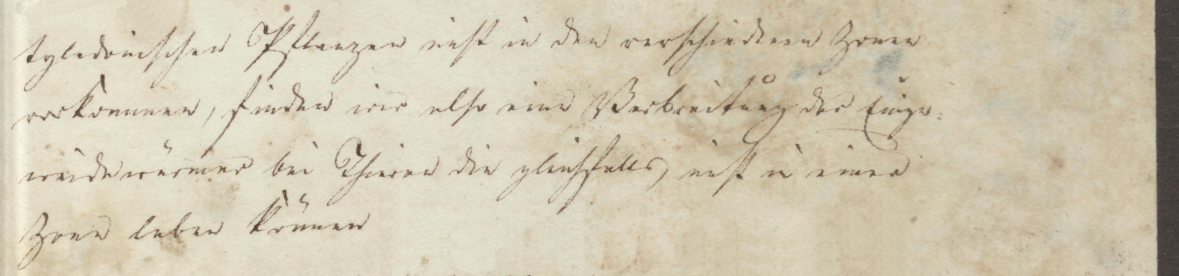} &
    \includegraphics[width=0.30 \linewidth, height=20mm]{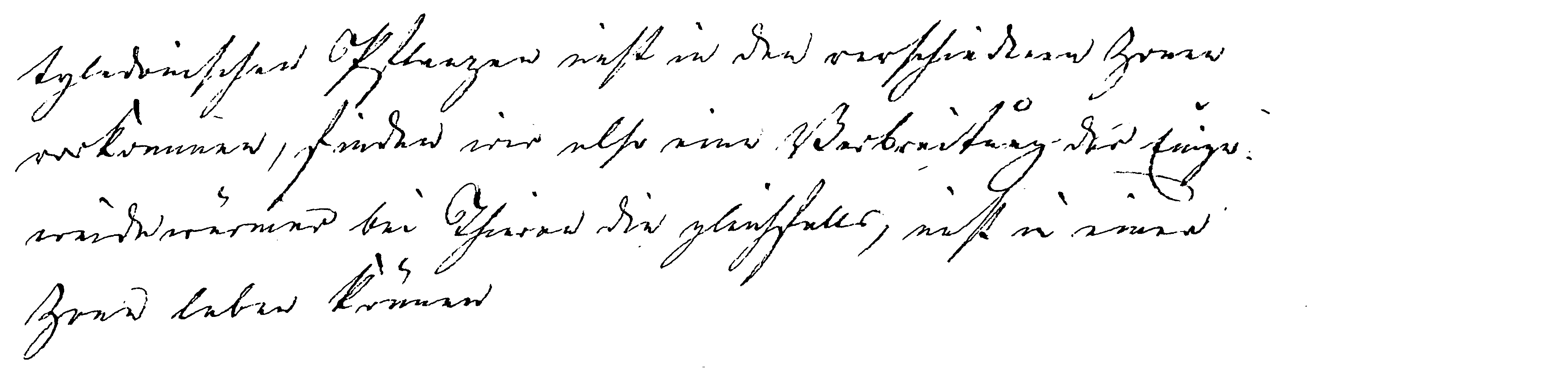}&
    \includegraphics[width=0.30 \linewidth, height=20mm]{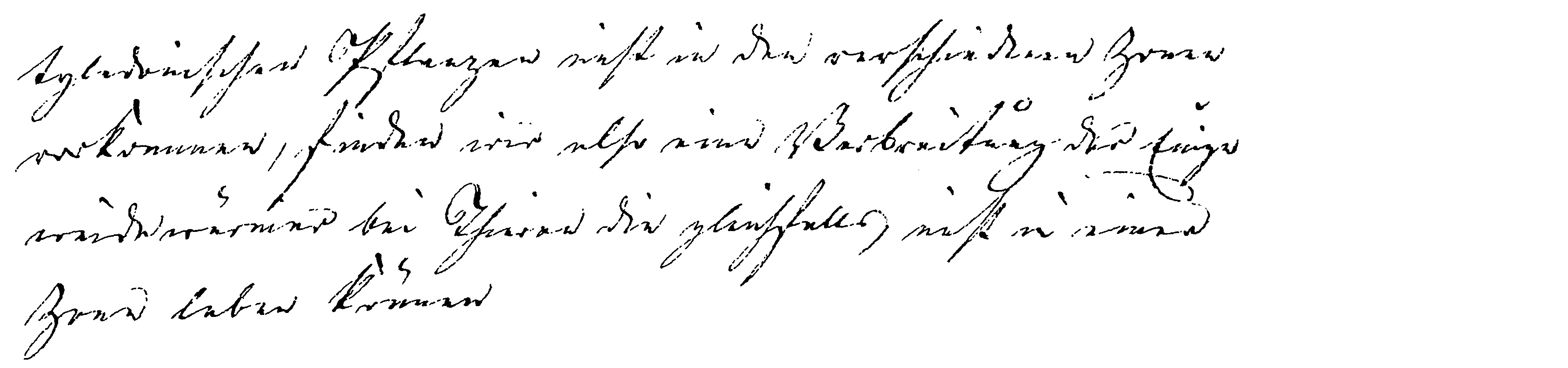}
    \\ \noalign{\smallskip}   
    
 \end{tabular}

 \caption{\textbf{Qualitative analysis of Text-DIAE in binarization task.} Images in columns are: Left: Original image, Middle: Binarized image using our proposed method., Right: Ground Truth image}
 \label{fig:our_results_dibcos_details}
% \end{center}
\end{figure*}

\begin{figure*}[t!]
  \begin{center}
  \footnotesize
  \renewcommand{\arraystretch}{0.5} 
  \begin{tabular}{cc}
 
    \includegraphics[width=0.4 \linewidth, height=40mm]{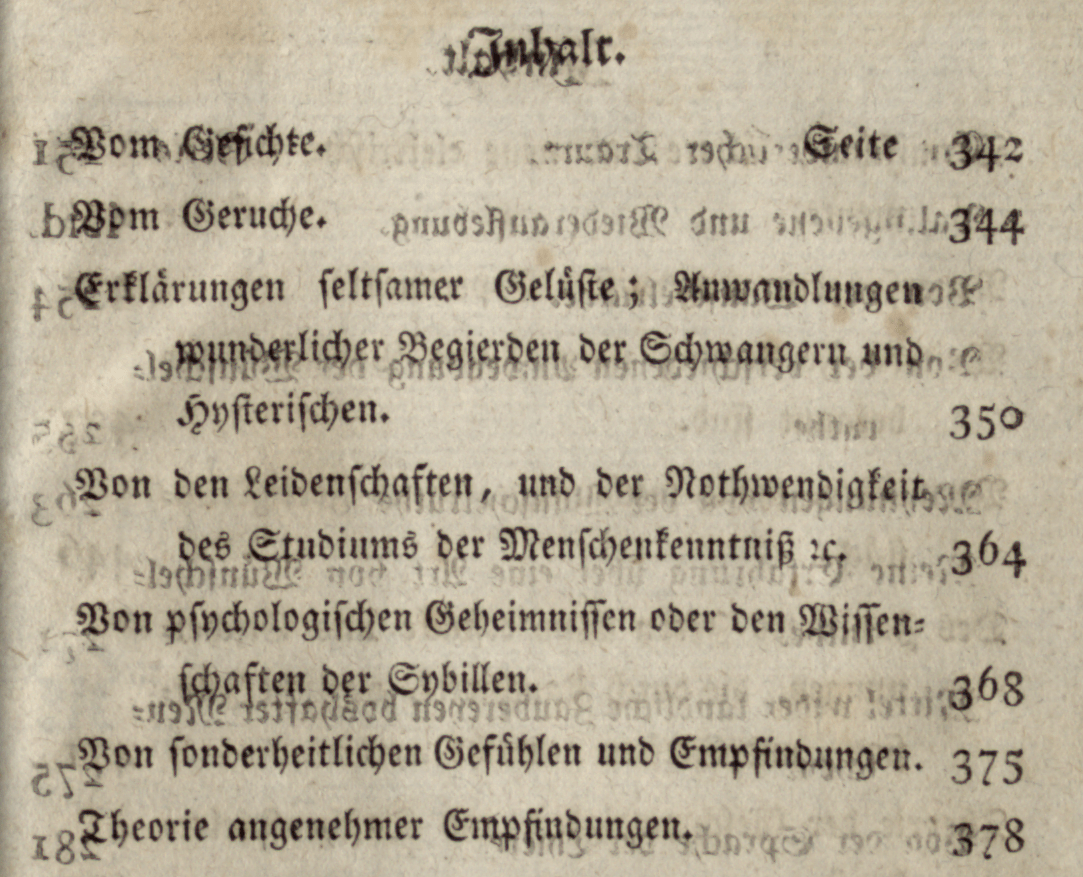} &
    \includegraphics[width=0.4 \linewidth, height=40mm]{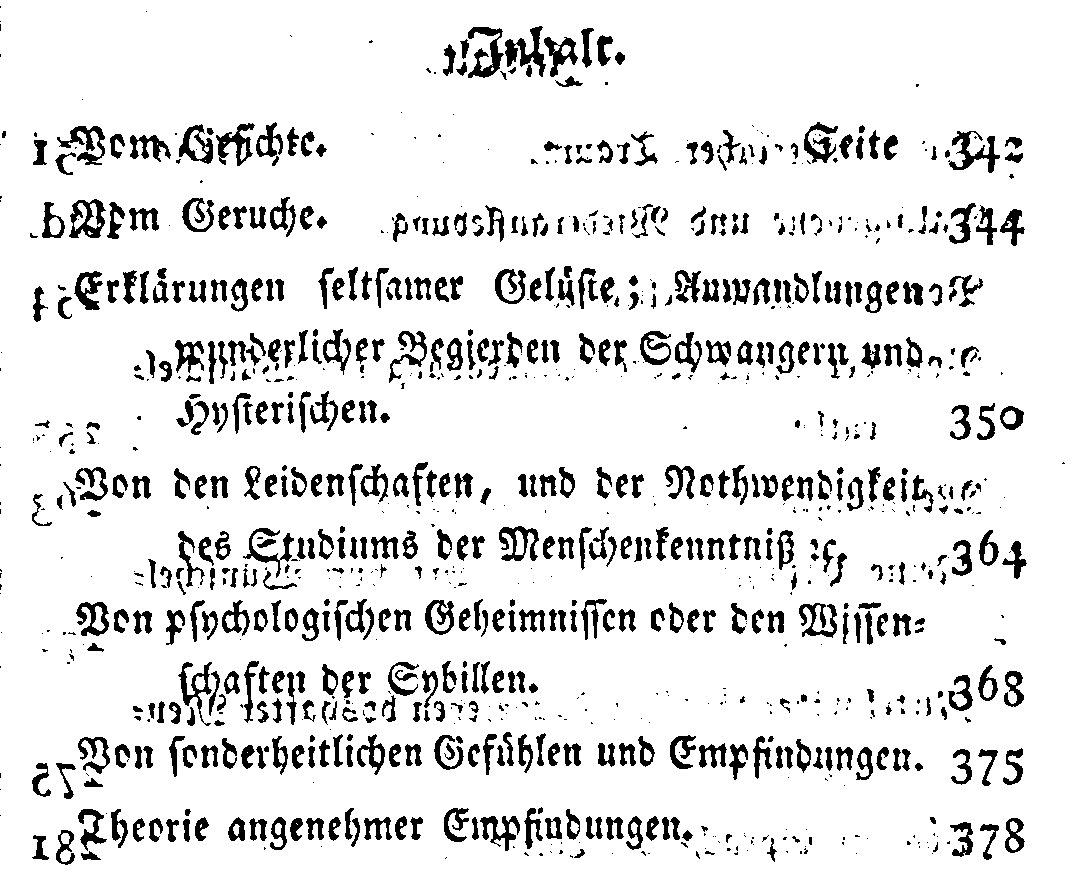}\\\noalign{\smallskip} 
    Original  & \cite{sauvola2000adaptive}\\\noalign{\smallskip} 
    \includegraphics[width=0.4 \linewidth, height=40mm]{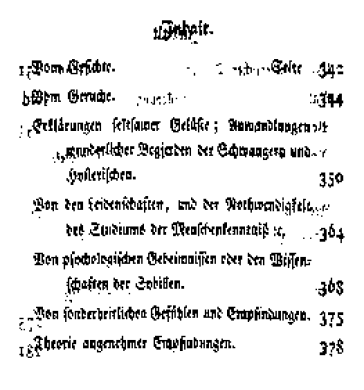}& 
    \includegraphics[width=0.4 \linewidth, height=40mm]{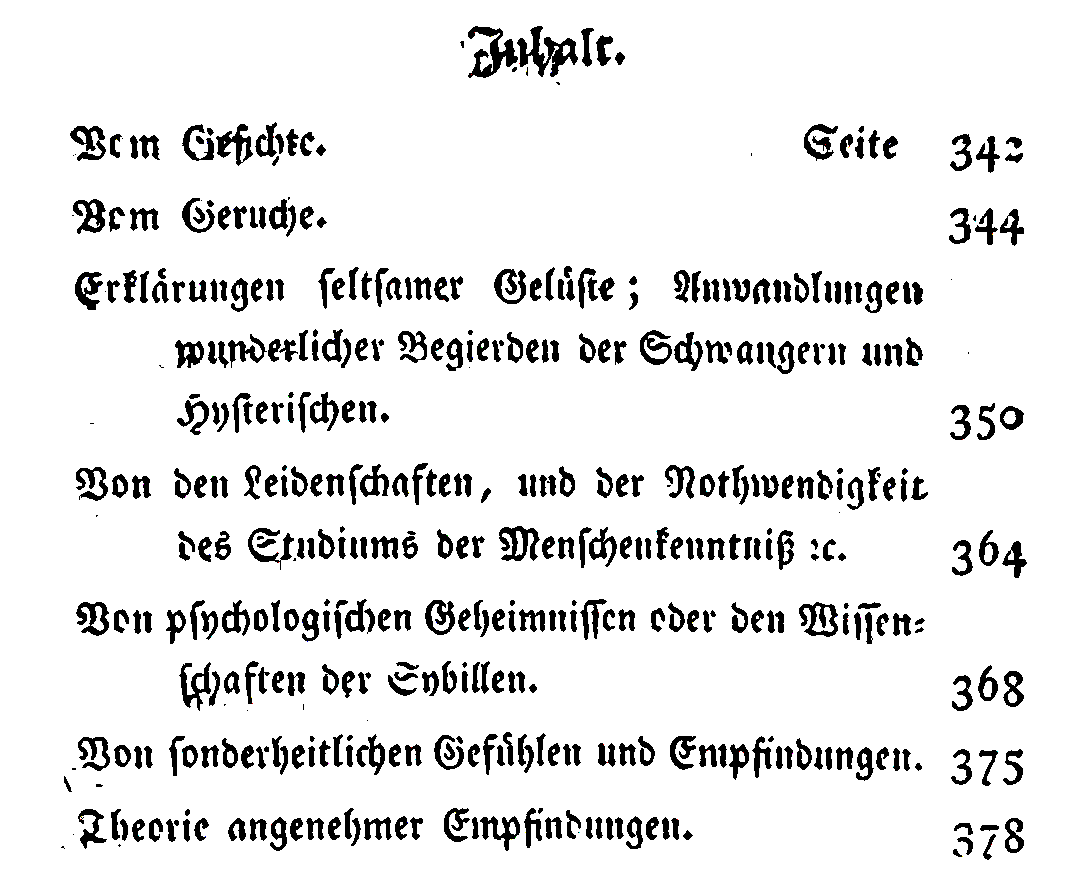}
    \\\noalign{\smallskip} 
    \cite{kang2021complex}& \cite{souibgui2022docentr} \\\noalign{\smallskip} 
    % Original & \cite{souibgui2022docentr} \\
    % \vspace{-0.2cm}
    % \\ \noalign{\smallskip}     
    \includegraphics[width=0.4 \linewidth, height=40mm]{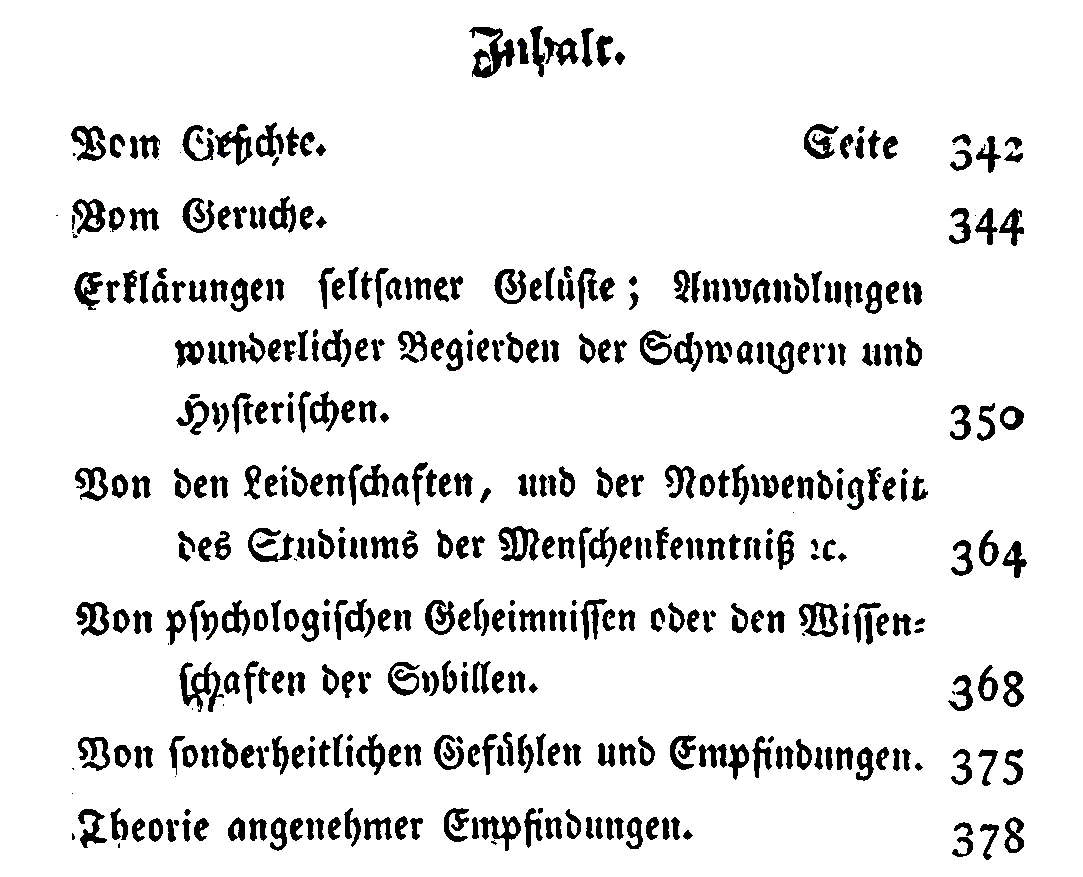}&
    \includegraphics[width=0.4 \linewidth, height=40mm]{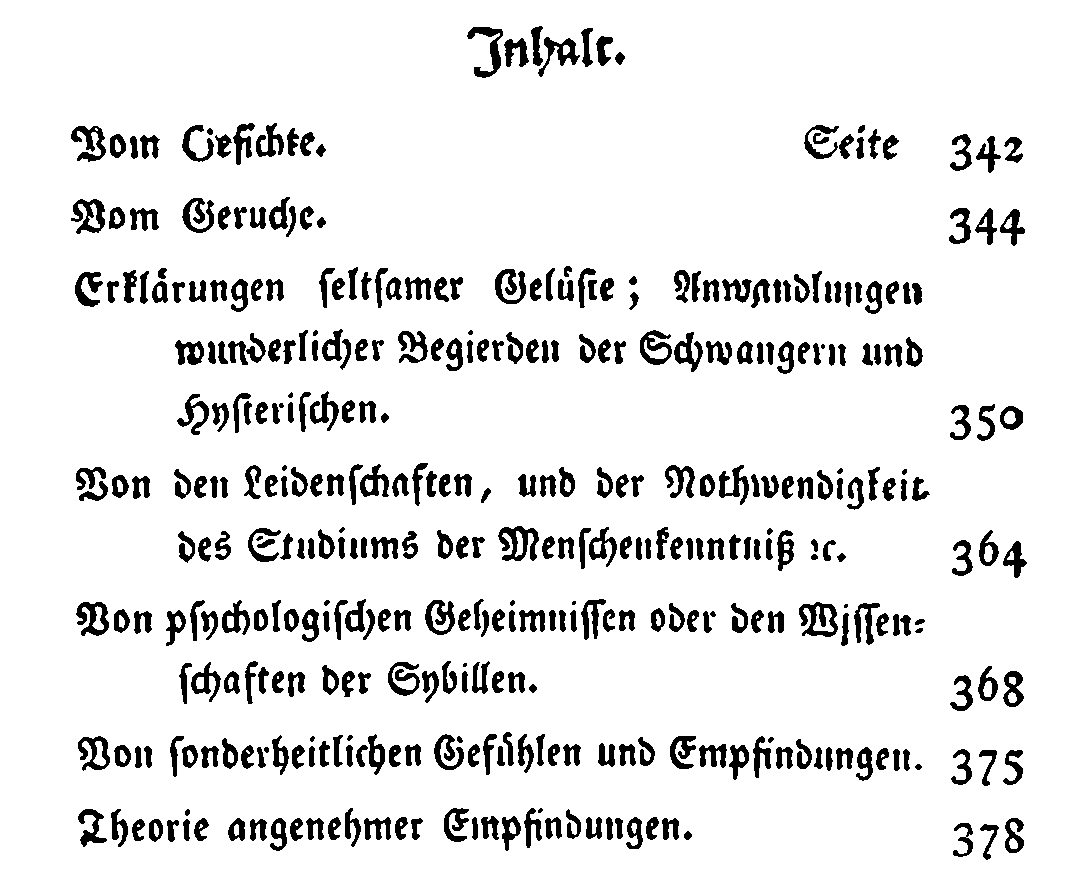}
    
    \\\noalign{\smallskip}

     \textbf{Ours} & Ground Truth
    \\%\noalign{\smallskip}   

    \end{tabular}
  \end{center}
  \caption{\textbf{Qualitative results of binarized samples:} We show the results of Text-DIAE on the document image binarization task. Given a degraded input example from DIBCO 2017, our model performs significantly better qualitatively compared to previous approaches.}
  \label{fig:binarization_qualitative_2}
%   \vspace{-0.5cm}
\end{figure*}

\noindent
\textbf{Text-DIAE recovers show-through.} The show-through problem appears when ink impressions from one side of the paper start appearing on the other side, making the document almost illegible. As shown in the results of examples (2nd, 3rd, 4th, 5th and 6th) of Figure~\ref{fig:our_results_dibcos_details} Text-DIAE binarized results get rid of the show-through successfully to make the document easily readable. An other qualitative comparison with other approaches (refer to Figure~\ref{fig:binarization_qualitative_2}) reveals the superiority of our approach than the related work in this task. 
\\

\noindent
\textbf{Text-DIAE recovers smears or stains.} As shown in the 1st example of Figure~\ref{fig:our_results_dibcos_details} documents also suffer from stains or smears and need to be recovered for proper readability. Text-DIAE not only successfully recovers the text (foreground) but also smoothens its boundary pixels.  
\\

\noindent
\textbf{Text-DIAE recovers faint characters or weak text.} Every individual glyph (character) inside a document can either appear faded due to the ink quality or paint as they start shrinking with time. Even adaptive handcrafted binarization approaches fail to recover and extract the text accurately. We illustrate in Figure~\ref{fig:ocr_dibco} an example of thin parchment of text and run the Tesseract-OCR engine (with their already in-built adaptive binarization tool) for validating the Text-DIAE model. The predicted OCR output from the original input image is compared with the binarized results from Text-DIAE, DocEnTr and the ground-truth image. The improvement in CER when compared with the original input and DocEnTr proves that Text-DIAE works substantially well to recover weak text for an OCR to read. Also, in the 8th example of Figure~\ref{fig:our_results_dibcos_details}, Text-DIAE performs quite good for thin characters under a faded background.  
\\

\noindent
\textbf{Text-DIAE recovers contrast variations.} Most of the times in historical degraded documents, they suffer from huge differences between high/low pixels due to occlusion, illumination and other noisy factors. In the examples shown in Figure~\ref{fig:our_results_dibcos_details} the binarized results have been obtained under several different contrasting background variations. This can be attributed to the feature representations learnt during pre-training stage by Text-DIAE by applying background noise transformation.
\\

\begin{figure*}[t]
% \begin{center}
\centering
  
 \begin{tabular}{ccc}
 
    \includegraphics[width=0.30 \linewidth, height=20mm]{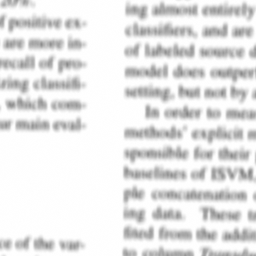} &
    \includegraphics[width=0.30 \linewidth, height=20mm]{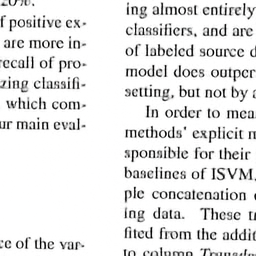}&
    \includegraphics[width=0.30 \linewidth, height=20mm]{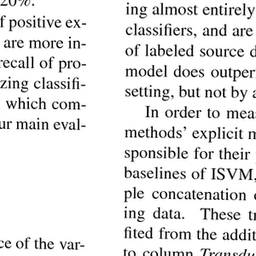}
    \\ \noalign{\smallskip} 
    
    \includegraphics[width=0.30 \linewidth, height=20mm]{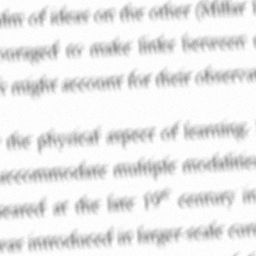} &
    \includegraphics[width=0.30 \linewidth, height=20mm]{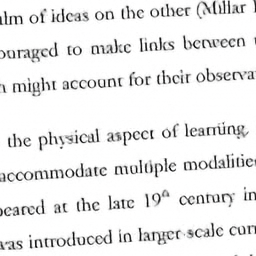}&
    \includegraphics[width=0.30 \linewidth, height=20mm]{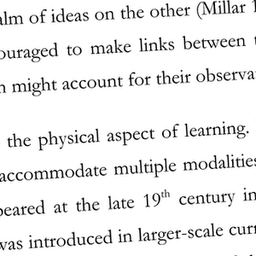}
    \\ \noalign{\smallskip}    
    
    \includegraphics[width=0.30 \linewidth, height=20mm]{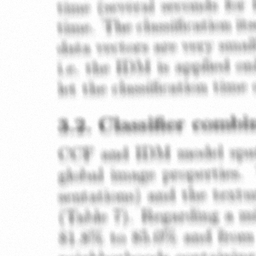} &
    \includegraphics[width=0.30 \linewidth, height=20mm]{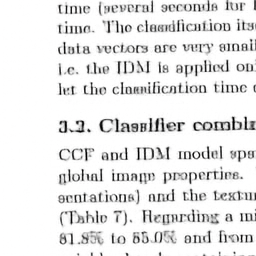}&
    \includegraphics[width=0.30 \linewidth, height=20mm]{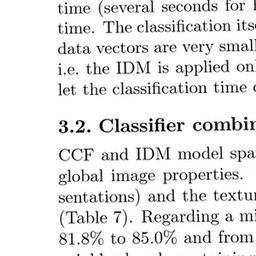}
    \\ \noalign{\smallskip}  
    
    \includegraphics[width=0.30 \linewidth, height=20mm]{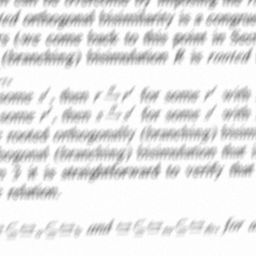} &
    \includegraphics[width=0.30 \linewidth, height=20mm]{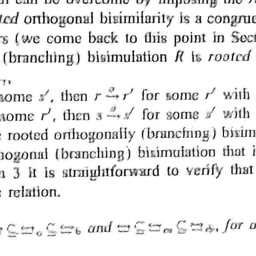}&
    \includegraphics[width=0.30 \linewidth, height=20mm]{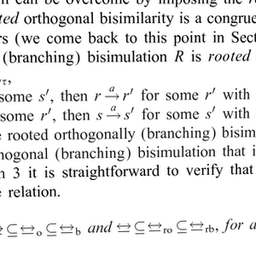}
    \\ \noalign{\smallskip}  
    
    \includegraphics[width=0.30 \linewidth, height=20mm]{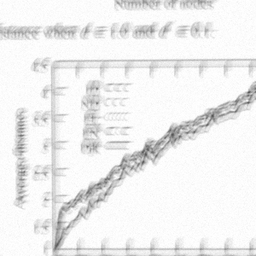} &
    \includegraphics[width=0.30 \linewidth, height=20mm]{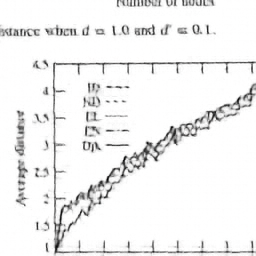}&
    \includegraphics[width=0.30 \linewidth, height=20mm]{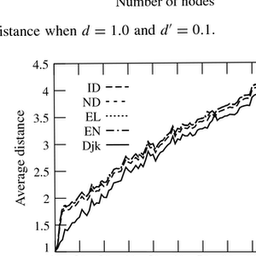}
    \\ \noalign{\smallskip}  
    
    \includegraphics[width=0.30 \linewidth, height=20mm]{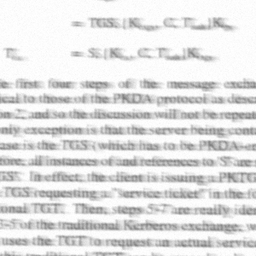} &
    \includegraphics[width=0.30 \linewidth, height=20mm]{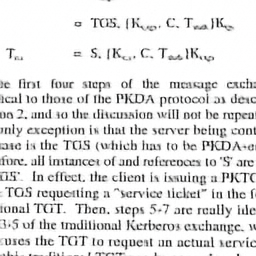}&
    \includegraphics[width=0.30 \linewidth, height=20mm]{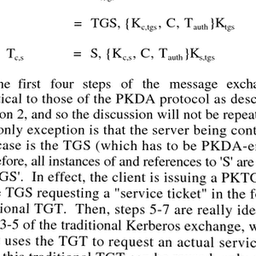}
    \\ \noalign{\smallskip}

    \includegraphics[width=0.30 \linewidth, height=20mm]{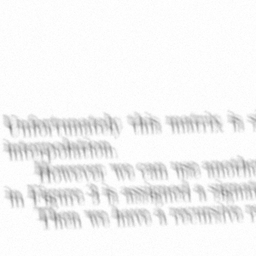} &
    \includegraphics[width=0.30 \linewidth, height=20mm]{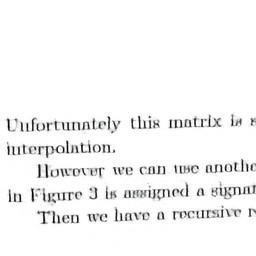}&
    \includegraphics[width=0.30 \linewidth, height=20mm]{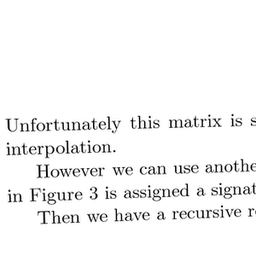}
    \\ \noalign{\smallskip}     
    
    \includegraphics[width=0.30 \linewidth, height=20mm]{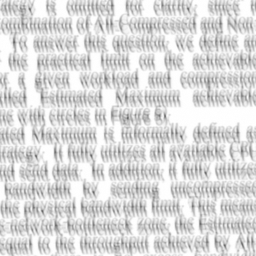} &
    \includegraphics[width=0.30 \linewidth, height=20mm]{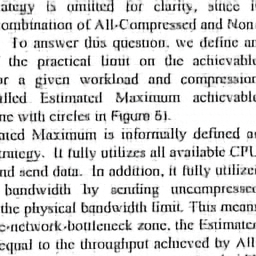}&
    \includegraphics[width=0.30 \linewidth, height=20mm]{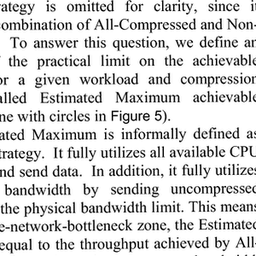}
    \\ \noalign{\smallskip}   
    
 \end{tabular}

 \caption{\textbf{Qualitative analysis of Text-DIAE in deblurring task.} Images in columns are: Left: Original image, Middle: Deblurred image output from Text-DIAE, Right: Ground Truth image}
 \label{fig:our_results_deblur_details}
% \end{center}
\end{figure*}

\noindent
\textbf{Text-DIAE excels in deblurring task.} When it comes to camera-captured documents, it often suffers from different kinds of blurring. Motion blur artefacts occur due to the relative speed between the object and the camera, where a sudden camera movement leads to a degradation in the captured image. The out-of-focus blur mainly occur when light fails to converge in the during camera capture. In this work, Text-DIAE was mainly applied to the benchmark data proposed in~\cite{hradivs2015convolutional} to see how the model performs in document deblurring. The results shown in Figure~\ref{fig:our_results_deblur_details} exhibit how the deblurred results from Text-DIAE achieves almost ground-truth level recovery from both motion-blur and out-of-focus blurring effects. The enhancement of the blurred images also help to capture the text from the deblurred sample, an illustration of which has been already presented in the main section of the paper. Also, the deblurring learning objective which is applied during the pre-training stage contributes significantly to improve the performance on the blurred document samples.    
\\

\noindent
\textbf{Robustness Evaluation of Text-DIAE.} We use the same model pre-trained on the DIBCO datasets (2009, 2010, 2013, 2014, 2015, 2016), Palm Leaf images~\cite{burie2016icfhr2016} and IAM~\cite{marti2002iam} handwriting database and use it on fine-tuning for two different document image enhancement tasks: binarization and deblurring, without any special change in settings. This proves the robustness of Text-DIAE as it learns universal feature representations from different kinds of degradation (background noise, blur and masking) beneficial for multiple downstream tasks.

% \subsection{More ablations with Text-DIAE:}

%%%  ####

